%% file: main.tex
\documentclass[letterpaper]{article} % DO NOT CHANGE THIS
\usepackage{aaai24}  % DO NOT CHANGE THIS
\usepackage{times}  % DO NOT CHANGE THIS
\usepackage{helvet}  % DO NOT CHANGE THIS
\usepackage{courier}  % DO NOT CHANGE THIS
\usepackage[hyphens]{url}  % DO NOT CHANGE THIS
\usepackage{graphicx} % DO NOT CHANGE THIS
\usepackage{natbib}  % DO NOT CHANGE THIS AND DO NOT ADD ANY OPTIONS TO IT
\usepackage{caption} % DO NOT CHANGE THIS AND DO NOT ADD ANY OPTIONS TO IT
\usepackage{algorithm}

\usepackage{graphicx}
\usepackage{amsmath}
\usepackage{amssymb}
\usepackage{booktabs}
\usepackage{algorithm}
\usepackage{algpseudocode}
\usepackage{multirow}
\usepackage{enumitem}

\usepackage{times}
\usepackage{epsfig}
\usepackage{graphicx}
\usepackage{amsmath}
\usepackage{amssymb}
\usepackage{subcaption}
\usepackage{xcolor}

\usepackage{algorithmicx}
\usepackage{hyperref}

\usepackage{newfloat}
\usepackage{listings}
\DeclareCaptionStyle{ruled}{labelfont=normalfont,labelsep=colon,strut=off} % DO NOT CHANGE THIS
\floatstyle{ruled}
\newfloat{listing}{tb}{lst}{}
\floatname{listing}{Listing}

\title{Find the Lady: Permutation and Re-Synchronization of Deep Neural Networks}
\author{
    Carl De Sousa Trias\textsuperscript{\rm 1}, 
    Mihai Petru Mitrea\textsuperscript{\rm 1},
    Attilio Fiandrotti\textsuperscript{\rm 2},\\
    Marco Cagnazzo\textsuperscript{\rm 3},
    Sumanta Chaudhuri\textsuperscript{\rm 4},
    Enzo Tartaglione\textsuperscript{\rm 4}
}
\affiliations{
    \textsuperscript{\rm 1}T\'el\'ecom SudParis, Institut Polytechnique de Paris, France\\
    \textsuperscript{\rm 2}University of Turin, Italy\\
    \textsuperscript{\rm 3}University of Padua, Italy\\
    \textsuperscript{\rm 4} LTCI, T\'el\'ecom Paris, Institut Polytechnique de Paris, France\\
    \texttt{carl.de-sousa-trias@telecom-sudparis.eu}
}

\begin{document}

\maketitle

\begin{abstract}
Deep neural networks are characterized by multiple symmetrical, equi-loss solutions that are redundant. 
Thus, the order of neurons in a layer and feature maps can be given arbitrary permutations, without affecting (or minimally affecting) their output. If we shuffle these neurons, or if we apply to them some perturbations (like fine-tuning) can we put them back in the original order i.e. re-synchronize?
Is there a possible corruption threat? Answering these questions is important for applications like neural network white-box watermarking for ownership tracking and integrity verification. \\
We advance a method to re-synchronize the order of permuted neurons. Our method is also effective if neurons are further altered by parameter pruning, quantization, and fine-tuning, showing robustness to integrity attacks. Additionally, we provide theoretical and practical evidence for the usual means to corrupt the integrity of the model, resulting in a solution to counter it. We test our approach on popular computer vision datasets and models, and we illustrate the threat and our countermeasure on a popular white-box watermarking method.
\end{abstract}

%%%%%%%%% BODY TEXT
\input{sections/1_introduction.tex}

\input{sections/2_permuting_neurons}
\input{sections/3_resync_neurons}
\input{sections/4_experiments.tex}
\input{sections/5_conclusion.tex}

\section*{Acknowledgements}

This work was funded in part by the Digicosme Labex through the transversal project NewEmma and by Hi!PARIS Center on Data Analytics and Artificial Intelligence.
\bibliography{main}

\input{supplementary}

\end{document}

%% file: sections/1_introduction.tex
\section{Introduction}

The deployment of deep neural networks for solving complex tasks became massive, for both industrial and end-user-oriented applications.
These tasks are instantiated in a huge variety of applications, e.g. autonomous driving cars.
In this context, neural networks are in charge of safety-critical operations such as forecasting other vehicles' trajectories, acting on commands to dodge pedestrians, etc. The interest in protecting the integrity and the intellectual property of such networks has steadily increased even for non-critical tasks, like ChatGPT content detection~\cite{uchida2017embedding,adi2018turning,li2021survey}. Some watermarking techniques already allow embedding signatures inside deep models~\cite{uchida2017embedding,chen2019deepmarks,tartaglione2021delving}, but these are designed to be robust against conventional attacks, including fine-tuning, pruning, or quantization, and assume the original location of the watermarked parameters remains unchanged.
\begin{figure}[ht]
\centering
\includegraphics[width=0.85\columnwidth]{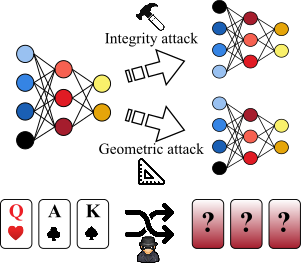}
\caption{Given some model (left), let us assume we permute the order of neurons and apply other types of corruption (right): are integrity checks at the neuron's level enough to verify the integrity of the model? And what about retrieving the signature in white-box watermarking? This problem resembles the ``find the lady/ three-card monte'' game, where the queen of hearts needs to be found out of shuffled cards.}
\label{fig:teaser}

\end{figure}
Neural networks, however, have internal symmetries such that entire neurons can be permuted, without impacting the overall computational graph. 
Once this happens, although the input-output function for the whole model does not change, the ordering of the parameters in the layer changes, and for instance, all the aforementioned watermarking approaches fail in retrieving the signature of the model, despite 
it still being there. This is referred to as \emph{geometric attack} in the multimedia watermarking community~\cite{wan2022comprehensive}, and we port the same concept to deep neural networks: the input-output relationship is preserved, but the order of the neurons is permuted, disallowing the recovery of signatures (Fig.~\ref{fig:teaser}). \\
Some studies \cite{hecht1990algebraic,ganju2018property,li2022fostering} already raised concerns about permutation in deep layers; yet, such a problem has not yet been studied in its general form, nor has its formal definition been stated. The first question we ask ourselves is whether the original ordering for the neurons can be retrieved, even when the applied permutation rule is lost. It is also a well-known fact that deep neural networks are redundant~\cite{setiono1997neural,agliari2020neural,wang2021convolutional} and some works enforce this towards improving the generalization capability of the neural network, like dropout~\cite{srivastava2014dropout}, while others detect such redundancies and prune them away~\cite{wang2021convolutional,chen2019drop,tartaglione2021serene}. Hence, it is not even clear whether it is possible to ``distinguish'', with no doubt, one neuron from all the others in the layer. This would be an important step to re-order (\emph{re-synchronize}) the neurons in the target layer. Besides, we ask the same question even in the case we apply some noise to the model: as a fact, the learning process for deep neural networks is noisy, and robustness towards the unequivocal identification of the parameters belonging to a neuron from the others in a noisy environment is important in the considered setup. \\
The main contributions of this study can be summarized as:
\begin{itemize}[noitemsep,nolistsep]
    \item we study the neuron redundancy case for deep neural networks, observing that despite some neurons showing the same input/output function, under the same input, their parameters can be consistently different;
    \item we explore different ways to re-synchronize a permuted model, showing and explaining fallacies for some of the most intuitive approaches;
    \item we put in  evidence a potential integrity threat for re-synchronized models and we highlight the counter-measure for it;
    \item we advance an effective solution to re-synchronize layers, even when subjected to noise, and we extensively validate it with four different noise sources, on five different datasets, and nine different architectures. 
\end{itemize}

%% file: sections/2_permuting_neurons.tex
\section{Permuting neurons}
\label{sec:perneuron}
\subsection{Preliminaries} 
\label{seq:perm}
\begin{figure*}
\centering
\begin{subfigure}[b]{0.24\textwidth}
    \includegraphics[width=\textwidth]{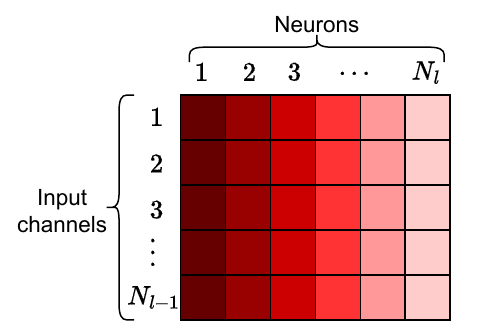}
    \caption{~}
    \label{fig:origlayl}
\end{subfigure}
%\hfill
\begin{subfigure}[b]{0.24\textwidth}
    \includegraphics[width=\textwidth]{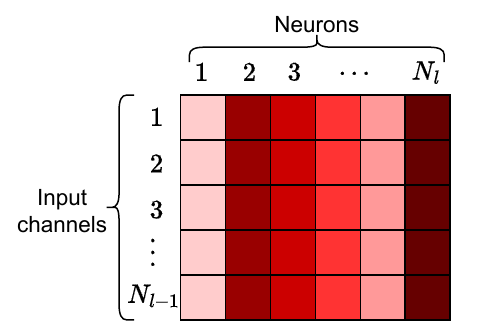}
    \caption{~}
    \label{fig:permlayl}
\end{subfigure}
\begin{subfigure}[b]{0.24\textwidth}
    \includegraphics[width=0.85\textwidth]{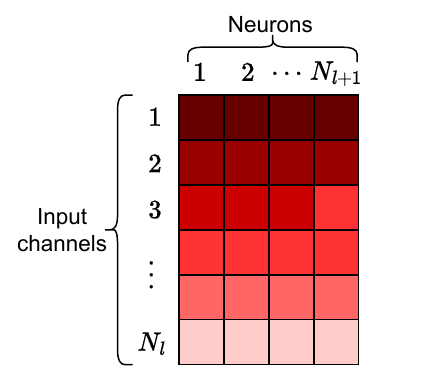}
    \caption{~}
    \label{fig:origlaylpone}
\end{subfigure}
%\hfill
\begin{subfigure}[b]{0.24\textwidth}
    \includegraphics[width=0.85\textwidth]{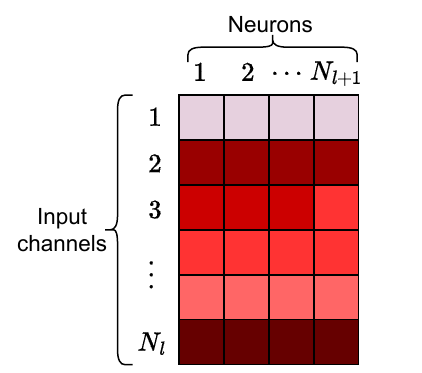}
    \caption{~}
    \label{fig:permlaylpone}
\end{subfigure}
\caption{Representation of the weights tensor for the $l$-th layer (a), permutation of neurons 1 and $N_l$ (b), representation of the weights tensor for layer $l+1$ (c) permutation on channels, following the same permutation of $l$ (d).}
\end{figure*}
In this section, we define the neuron permutation problem. For the sake of simplicity, we will exemplify the problem on a single fully connected layer without biases; however, the same conclusions hold for any other layer typology, e.g. convolutional or batch-normalization. Let us define the output $\boldsymbol{y}_{l}\in \mathbb{R}^{N_l\times 1}$ of the $l$-th layer:
\begin{equation}
    \label{eq:forwprop}
    \boldsymbol{y}_{l} = \varphi\left[\left< \boldsymbol{w}_{l}, \boldsymbol{y}_{l-1}\right> \right]
\end{equation} 
where $\boldsymbol{y}_{l-1}\in \mathbb{R}^{N_{l-1}\times 1}$ is the input, $\boldsymbol{w}_{l}\in \mathbb{R}^{N_{l-1}\times N_{l}}$ are the weights for the $l$-th layer (as displayed in Fig.~\ref{fig:origlayl}), $\left<\cdot\right>$ indicates inner product, and $\varphi(\cdot)$ is the activation function. Let us consider the case a permutation $\pi_l$ is applied on the neurons of the $l$-th layer; the elements in the permutation matrix $P_{\pi_l}\in \mathbb{R}^{N_{l}\times N_{l}}$ are:
\begin{equation}
   (P_{\pi_l})_{i, j} = \left\{
   \begin{array}{ll}
    1   & \text{if } j=\pi_l(i)\\
    0   & \text{otherwise}.
   \end{array}
   \right.
\end{equation}
The neurons are permuted, and the ordering for the input channels $\boldsymbol{y}_{l-1}$ remains intact (Fig.~\ref{fig:permlayl}). Hence, the permuted output for the $l$-th layer will be
\begin{equation}
    \boldsymbol{y}_{l}^{\pi_l} = \varphi\left( \left< \boldsymbol{w}_{l}^{\pi_l}, \boldsymbol{y}_{l-1} \right>\right),
\end{equation} 
\begin{equation}
\label{eq:neuperm}
    \boldsymbol{w}_{l,c,-}^{\pi_l} = \left< P_{\pi_l}, \left(\boldsymbol{w}_{l,c,-} \right)\right>~\forall c;
\end{equation}
where $\boldsymbol{w}_{l,c,-}$ represents all elements of the $l$-th layer for the $c$-th channel, and $\boldsymbol{w}_{l,-,n}$ represents all elements of the $l$-th layer for the $n$-th neurons. After having applied $\pi_l$ at layer $l$, the output of the model is likely to be altered, as the propagated $\boldsymbol{y}_l^{\pi_l}\neq \boldsymbol{y}_l$, which is processed as input by the next layer (Fig.~\ref{fig:origlaylpone}). Hence, to maintain the output of the full model unaltered, we need to also permute the weights in layer $l+1$ 
\begin{equation}
\label{eq:chanperm}
    \boldsymbol{w}_{l+1,-,n}^{\pi_l} = \left< P_{\pi_l}, \left(\boldsymbol{w}_{l+1,-,n} \right)\right>~\forall n.
\end{equation}
In this way, the permuted outputs in the $l$-th layer will be correctly weighted in the next layer, and the neural network output will be unchanged (Fig.~\ref{fig:permlaylpone}). To illustrate our study, we define a companion dataset and an architecture, namely the CIFAR-10 and VGG-16 (without batch normalization), respectively. The model we will use as a reference is trained for $50$ epochs using SGD, with a learning rate $10^{-2}$, weight decay $10^{-4}$, and momentum $0.9$. Let the first convolutional layer of the fourth block of convolutions (where every block is separated by a maxpool layer) be our $l$-th layer.

\begin{figure}[t]
\centering
\includegraphics[width=0.9\columnwidth]{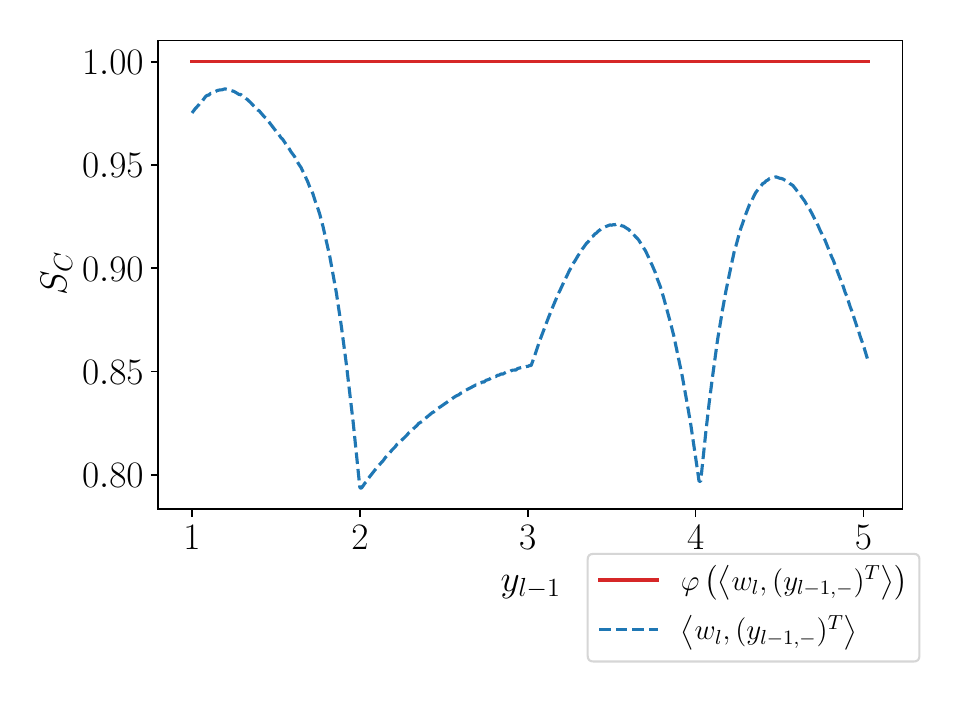}
\caption{Evolution of cosine similarity of two non-zero neurons before and after activation function for $y_{l-1}$ inputs.}
\label{fig:hope}
\end{figure}
\subsection{Any hope to recover the original order?} 
\label{sec:hope}
Assuming the $P_{\pi_l}\in \mathbb{R}^{N_{l}\times N_{l}}$ matrix is known, the answer is straightforward. Yet, the question becomes hard to answer when the $P_{\pi_l}\in \mathbb{R}^{N_{l}\times N_{l}}$ matrix is unknown. The difficulty derives from the fact that neural network models internally have many redundancies~\cite{setiono1997neural,wang2021convolutional} that can a priori cast confusion when trying to find the initial order. Many approaches, like dropout~\cite{srivastava2014dropout}, enforce this to make deep models robust against noise. Consider the case in which two neurons belonging to the $l$-th layer are \emph{redundant}, and let them be denoted by the $i$-th and the $j$-th, with the parameters $\boldsymbol{w}_{l,-,i}$ and $\boldsymbol{w}_{l,-,j}$. Given some $\xi$-th sample in $\mathcal{D}$, with $\mathcal{D}$ being the dataset the model is trained on, $y_{l,i}^\xi = y_{l,j}^\xi$. From this, we can have two scenarios: 
\begin{itemize}[noitemsep,nolistsep]
    \item $\boldsymbol{w}_{l,-,i} = \boldsymbol{w}_{l,-,j}$: in this case, the $i$-th and the $j$-th neuron share exactly the same parameters. As such, since they receive the same input $\boldsymbol{y}_{l-1}^\xi$, and by construction, they have all the same activation function $\varphi(\cdot)$, they map the same function and they are, hence, identical. Since they are the same from all points of view, ordering them one way or another does not matter. 
    \item $\boldsymbol{w}_{l,-,i} \neq \boldsymbol{w}_{l,-,j}$: in this case, the $i$-th and the $j$-th neuron have a different set of parameters, but share the same outputs for some samples in $\mathcal{D}$. 
\end{itemize}
The second case is the most interesting: is it possible to recover the original ordering of neurons exhibiting the same output under the same input? It is easy to prove that 
\begin{equation}
    \label{eq:cond}
    \boldsymbol{w}_{l,-,i} = k \cdot \boldsymbol{w}_{l,-,j} \Rightarrow y_{l,i}^\xi = k\cdot y_{l,j}^\xi \forall \xi,
\end{equation}    
with $k\in \mathbb{R}$ being some scalar quantity. To test whether two neurons are extracting the same information, we can compute the cosine similarity $S_C(y_{l,i}, y_{l,j})$ between their outputs, and ask that it is exactly one: from this, we obtain
\begin{equation}
    \label{eq:condequ}
    \sum_\xi y_{l,i}^{\xi} y_{l,j}^{\xi} = \sqrt{\sum_\xi \left(y_{l,i}^{\xi}\right)^2}\sqrt{\sum_\xi \left(y_{l,j}^{\xi}\right)^2}.
\end{equation}
From \eqref{eq:forwprop} it is clear that, having non-linear activations and in general $N_{l-1} > N_{l}$, \eqref{eq:condequ} is satisfiable for $\boldsymbol{w}_{l,-,i}\neq \boldsymbol{w}_{l,-,j}$.\\
Let us observe this empirically, using our companion setup: we select 2 neurons $i$,$j$ of the $l$-th layer such that their cosine similarity $S_C(y_{l,i}, y_{l,j})=1$, for several values of $k$. Since $l$ is a convolutional layer, we know that $\boldsymbol{y}_{l,i}^{\xi}\in \mathbb{R}^{1\times M_{l}}$, where $M_{l}$ is a function of the input size for $l$, kernel size and stride. Hence, we are able here to plot the cosine similarity given the input of one single $\xi$-th sample and to track the change of the similarity between $\boldsymbol{y}_{l-1}^\xi$ and $\boldsymbol{y}_{l-1}^{\xi+1}$. Fig.~\ref{fig:hope} displays the cosine similarity between two neurons in the $l$-th layer before and after the activation function. Despite the cosine similarity remaining to one, this happens thanks to the non-linear activation, as the pre-activation potentials are less correlated. Furthermore, we observe that the parameters of these neurons are essentially de-correlated, as their cosine similarity values $-0.02$. 
This shows that even if two neurons have a similar (non-zero) response to the same input, their internal function (before the non-linearity) can be different.  
This gives us hope to distinguish each neuron, hence, retrieving the original ordering of the neurons.

%% file: sections/3_resync_neurons.tex
\section{Re-synchronizing neurons}
\label{sec:resync}
In this section, we first define against which additional modifications, applied in conjunction with the permutation, the counterattack should still retrieve the original order, namely: Gaussian noise, fine-tuning, pruning, and quantization. Second, we explore the potential counterattack solutions by presenting methods of the state-of-the-art and showing where they worked and failed. Finally, we present our method leveraging the cosine similarity to recover the original order.
\subsection{Robustness in retrieving the original order}
\label{sec:modif}
In the previous section, we discussed how neurons can be permuted inside a neural network without impacting the model performance. In this section, assuming the initial permutation matrix is no longer available, we will explore
ways to recover the original ordering for permuted neurons, even when they are possibly modified. In particular, we will explore robustness in retrieving the original order when undergoing four different transformations:

\begin{itemize}[noitemsep,nolistsep]
    \item \textbf{Gaussian noise addition}: we apply an additive noise $\mathcal{N}(0,\sigma_{l}\Omega)$, with $\Omega\geq 0$, $\sigma_{l}$ standard deviation of $l$.
    \item \textbf{fine-tuning}: we resume the original training of the model with $\Theta$ standing for the ratio of fine-tuning epochs to the original training epochs.    
    \item \textbf{quantization}: we reduce the number of bits $B$ used to represent the parameters of the model.
    \item \textbf{magnitude pruning}: we mask the $T$ fraction of the smallest weights of the model, according to the $\ell_1$-norm. 
\end{itemize}
Even when the model undergoes these transformations, our goal is to be able to recover the original ordering for the model: we denote by $\Psi$ as the fraction of neurons we were able to place back to their original position (multiplied by $100$), and we shall refer it as \emph{re-synchronization success rate}. Here follows a sequence of approaches aiming at bringing $\Psi$ close to $100$, under the aforementioned transformations.

\subsection{In the search of the lost synchronization}
\label{sec:inthesearch}
The next sections explore the different methods to solve the permutation problem.

\paragraph{Finding the canonical space: rank the neurons} Our first approach consists of ranking all the neurons in the $l$-th layer according to some specific scoring function. For instance, we can attempt to look at the intrinsic properties of the neurons inside the layer, like their weight norm, to perform a ranking \cite{ganju2018property}.
\begin{figure*}
\centering
\begin{subfigure}[ht]{0.47\textwidth}
    \includegraphics[width=\textwidth]{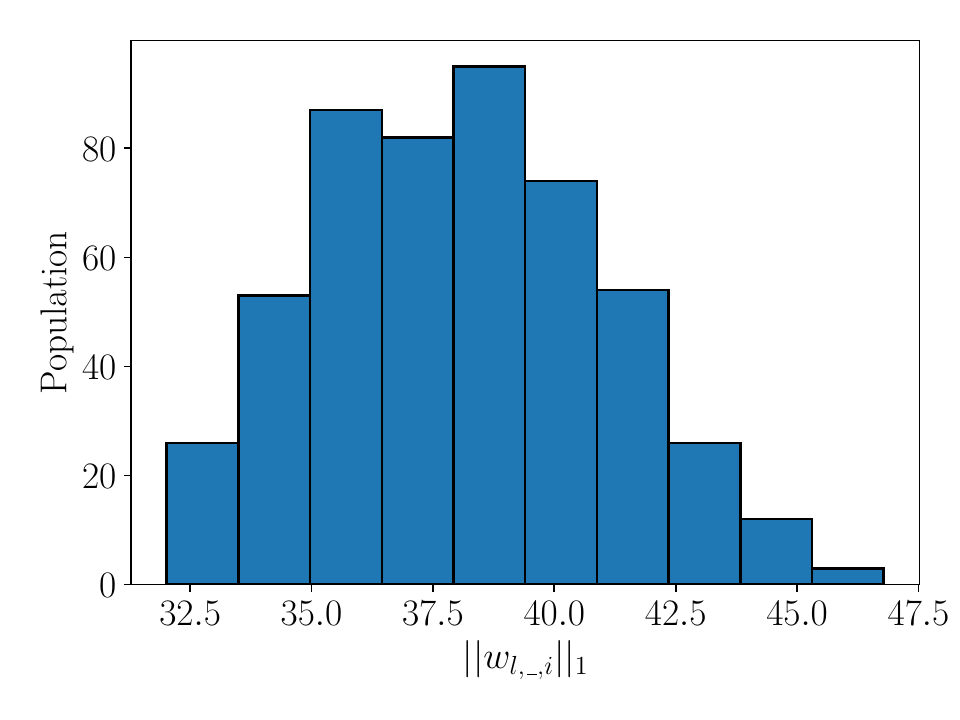}
    \caption{}
    \label{fig:method0_dist}
\end{subfigure}
\begin{subfigure}[ht]{0.47\textwidth}
    \includegraphics[width=\textwidth]{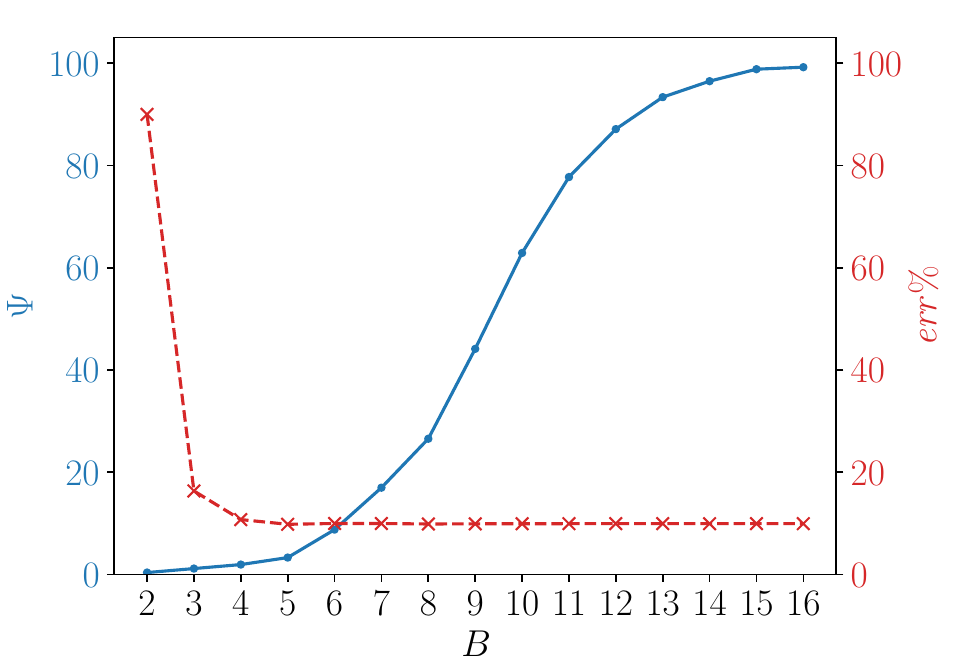}
    \caption{}
    \label{fig:method0_rob}
\end{subfigure}
\caption{ (a) L1 norm distribution of the neurons of the $l$-th layer a VGG-16 model trained on CIFAR10. (b) Robustness of ranking L1 norms of neurons, against quantization. $\Psi$ is on the left axis in blue and the $\textit{err \%}$ on the right axis in red.}
\label{fig:method0}
\end{figure*}
Unfortunately, this approach is not general: there are specific cases, like spherical neurons~\cite{lei2019octree} in which the parameters are normalized and, for instance, not possible to be ranked according to their norm. This effect is not limited to these special models: if we plot the distribution of the norms for the $l$-th layer in our companion VGG-16 model trained, as represented in Fig.~\ref{fig:method0_dist}, we observe that typically the values for the norm of the neuron's parameters are in a very small domain: for instance, the minimal gap between these norms is in the order of $10^{-5}$. 
We expect, hence, that this ranking is very sensitive to all the aforementioned transformations. As an example, Fig.~\ref{fig:method0_rob} displays the non-robustness against quantization attack: the neurons are permuted (blue line, the higher the better) before losing any performance on the task (red line, the lower the better).

\paragraph{Creating a trigger set}
A second approach could be to learn an input $\boldsymbol{y}_{l-1}$ such that the output $\boldsymbol{y}_{l}$ permits to identify the neurons. With our first approach, we aim to learn a $\boldsymbol{y}_{l-1}$ such that we maximize the distance between all the neurons' outputs. Then, we say the norm of the output corresponds to the ranking of the neuron itself. 
Empirically, in our companion setup, we observe that this approach is not robust to fine-tuning. Indeed, despite having a minimum gap between outputs larger than one, after just an extra $1\%$ of training, we observe a re-synchronization success rate already dropping to $10\%$, or in other words, we are not able to recover the exact position for the $90\%$ of neurons. This effect confirms our idea that neurons could have similar behavior for the same input which makes them easily swapped after any modifications. Another approach is developed in \cite{li2022fostering} which aims to learn a set of inputs to identify a neuron based on the response to the trigger set. However, this method seems ineffective since the re-synchronization success rate never reaches $100\%$ and it was only tested on the first layers of the neural network. \\
Finally, we simplify the problem by identifying each neuron independently from the others. If we can do so, then we will also be able to re-synchronize the whole layer. Towards this end, using a similar strategy heavily employed in many interpretability works~\cite{suzuki2017deep}, we can learn the input $\boldsymbol{y}_{l-1}$ which maximizes the response of the $i$-th neuron only, and at the same time minimizes the response of all the others. With this method we can create a set of $N_l$ inputs for the $l$-th layer, to identify all its neurons.\\
This approach shows its robustness to all the modifications, but has a big drawback: it demands a lot of memory to store the learned inputs (we need one input per neuron, hence the space complexity is $\mathcal{O}(N_{l-1}\cdot N_{l} \cdot M_{l-1})$, where $M_{l-1}$ is the size of each output coming from $l-1$). Besides, we need also a consistent computational effort, as we need to forward a batch of $N_{l}$ inputs. This makes the ``re-synchronizer'' overall bigger than the model itself and becomes prohibitive. 

\begin{algorithm}
\caption{Re-synchonization algorithm.}\label{alg:np}
\begin{algorithmic}
\State \textbf{Inputs:} the original model $\Gamma$, the altered model $\tilde{\Gamma}_{\pi_l}$, the number of layers of these models $L$.
\State \textbf{Output:} The re-synchronized model $\tilde{\Gamma}$
\For{$ l = \{1,\dotsc,L-1\}$} 
\State \textbf{Step 1:} Compute score metric on ${\tilde{\boldsymbol{w}}}_{l}^{\pi_l}$
\State $\boldsymbol{w}_{l}  \in \mathbb{R}^{N_{l-1}\times N_{l}} \gets $  parameters in $l^{th}$ layer of $\Gamma$ 
\State $\tilde{\boldsymbol{w}}^{\pi_l}_{l}  \in \mathbb{R}^{N_{l-1}\times N_{l}} \gets $   parameters in $l^{th}$ layer of $\tilde{\Gamma}_{\pi_l}$
\State $S \gets S_C(\boldsymbol{w}_{l},\tilde{\boldsymbol{w}}^{\pi_l}_{l}) = \frac{(\boldsymbol{w}_{l})^T \cdot \tilde{\boldsymbol{w}}^{\pi_l}_{l}}{\lVert \boldsymbol{w}_{l} \rVert _2 \lVert \tilde{\boldsymbol{w}}^{\pi_l}_{l} \rVert _2}$ 
\State \textbf{Step 2:} Obtain the permutation matrix $P_{\pi_l^{-1}}$
\State $P_{\pi_l^{-1}} \gets [0]_{N_l,N_l}$
\For{$ i = \{1,\dotsc,N_l\}$}
\State $j \gets \text{argmax}_i(S)$ 
\State $(P_{\pi_l^{-1}})_{i,j}=1$
\EndFor
\State \textbf{Step 3:} Permute neurons in $l^{th}$ layer of $\tilde{\Gamma}$ and channels 
\State in $(l+1)^{th}$ of $\tilde{\Gamma}_{\pi_l}$
\State $\tilde{\boldsymbol{w}}_{l} \gets \left< P_{\pi_l^{-1}}, \left(\tilde{\boldsymbol{w}}^{\pi_l}_{l,c,-} \right)\right>~\forall c$ \Comment{ equation (4)}
\State $\tilde{\boldsymbol{w}}_{l+1}^{\pi_{l}} \gets \left< P_{\pi_l^{-1}}, \left(\tilde{\boldsymbol{w}}^{\pi_l}_{l+1,-,n} \right)\right>~\forall n$ \Comment{ equation (5)}
\EndFor
\State \textbf{return} $\tilde{\Gamma}$ 
\end{algorithmic}
\end{algorithm}

\begin{figure}[t]
\centering
\includegraphics[width=0.97\columnwidth]{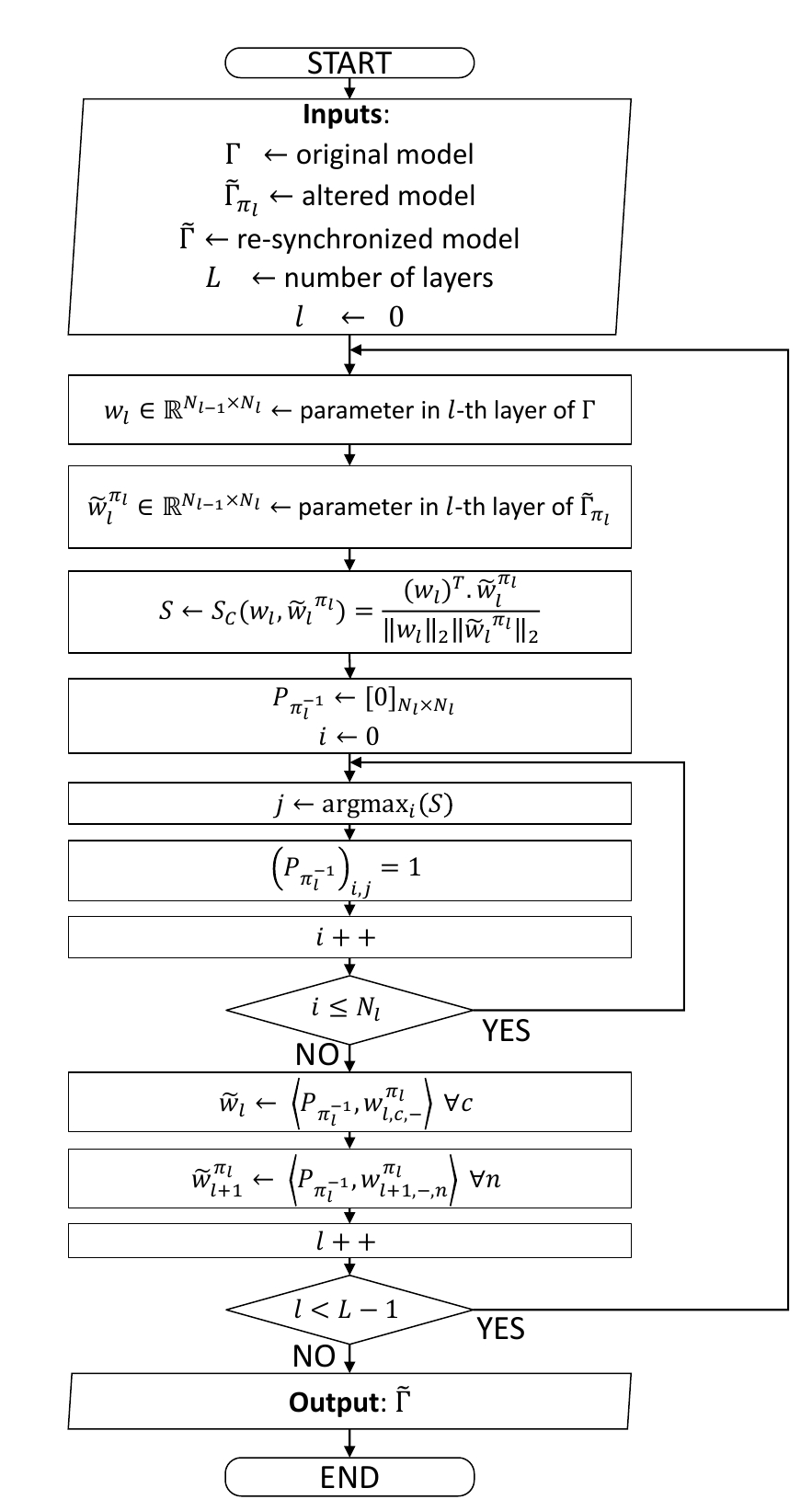}
\caption{Flowchart of Alg.~\ref{alg:np}.}
\label{fig:flo}
\end{figure}

\subsection{Find the lady by similarity}

\begin{figure*}[ht] 
\centering

\begin{subfigure}[b]{0.24\textwidth} 
    \includegraphics[width=\textwidth]{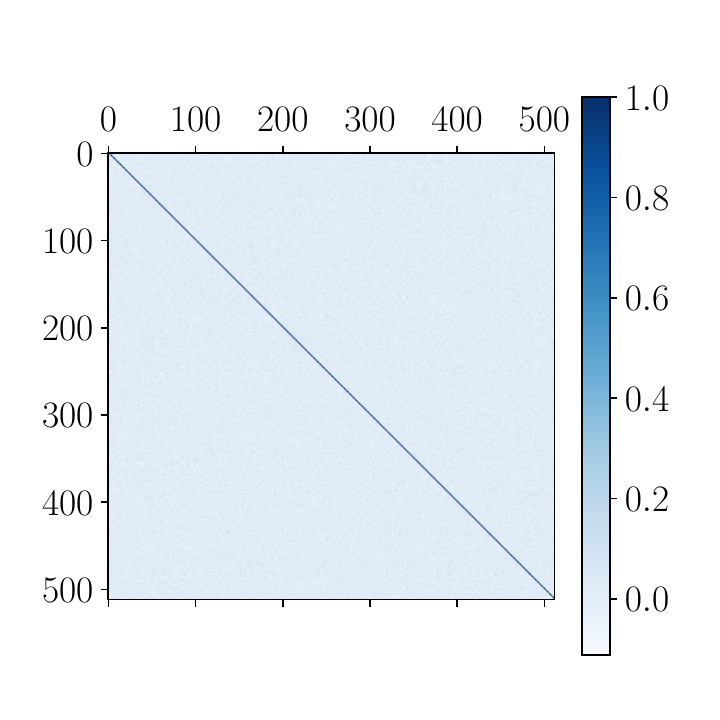}
    \caption{$S_C(\boldsymbol{w}_{l,i,-},{\boldsymbol{w}}_{l,i,-})$}
    \label{fig:cosmata}
\end{subfigure}
%\hfill
\begin{subfigure}[b]{0.24\textwidth} 
    \includegraphics[width=\textwidth]{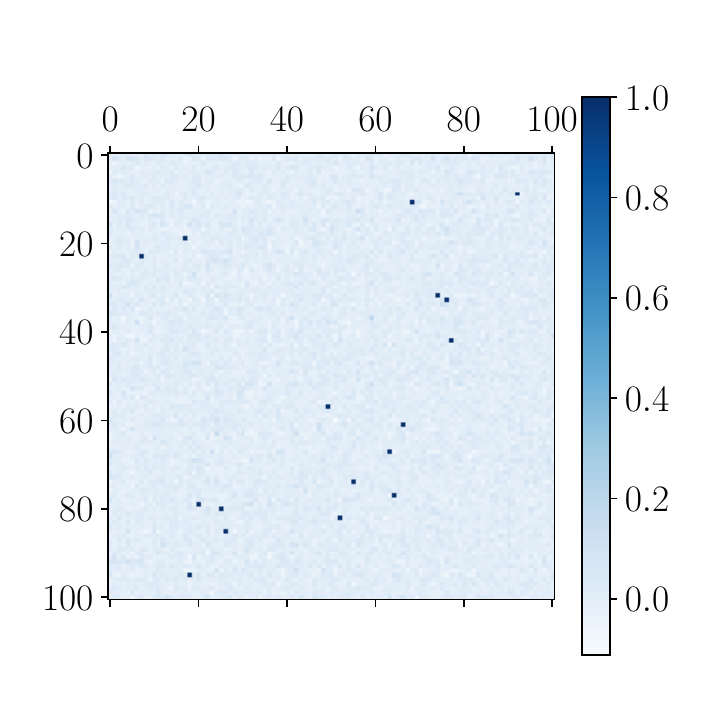}
    \caption{$S_C(\boldsymbol{w}_{l,i,-},{\boldsymbol{w}}_{l,i,-}^{\pi_l})$}
    \label{fig:cosmatb}
\end{subfigure}
\begin{subfigure}[b]{0.24\textwidth}
    \includegraphics[width=\textwidth]{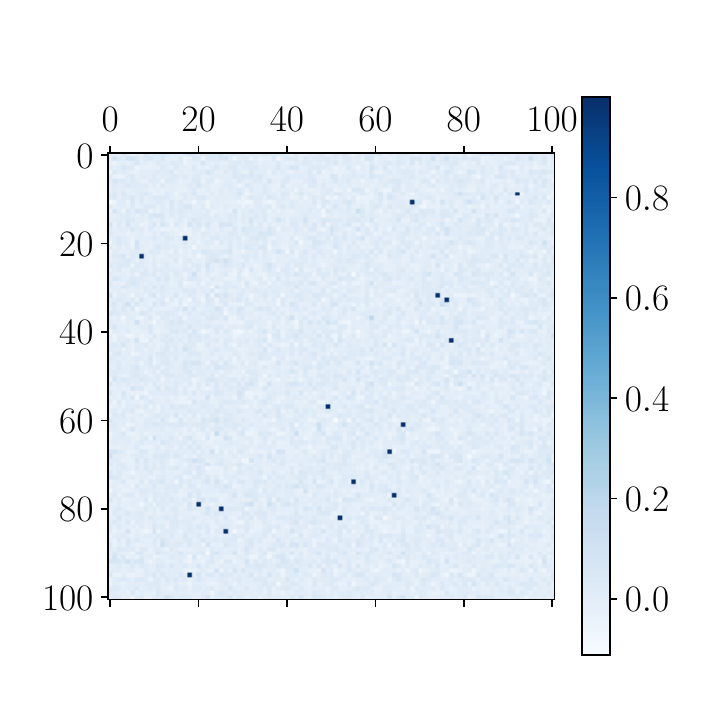}
    \caption{$S_C(\boldsymbol{w}_{l,i,-},\tilde{\boldsymbol{w}}_{l,i,-}^{\pi_l})$}
    \label{fig:cosmatc}
\end{subfigure}
\begin{subfigure}[b]{0.24\textwidth}
    \includegraphics[width=\textwidth]{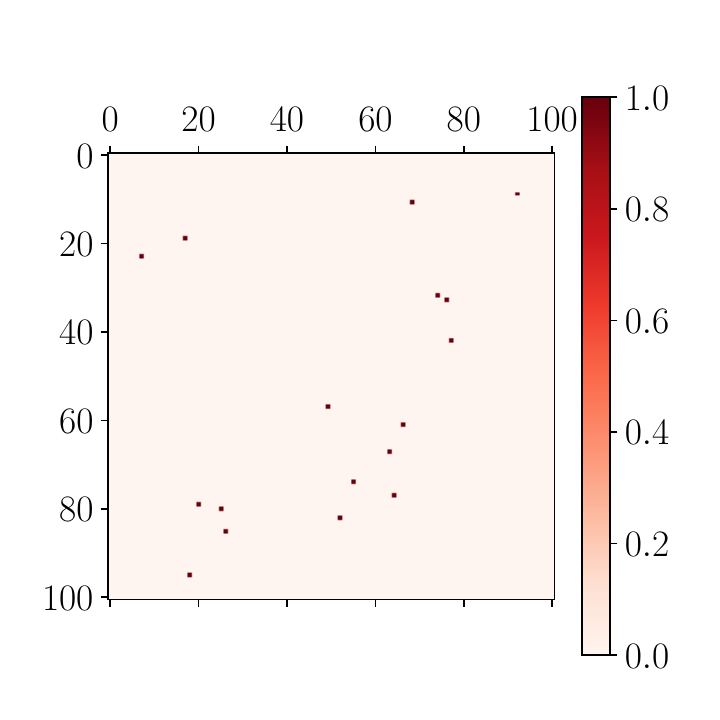}
    \caption{$P_{\pi_l}$}
    \label{fig:cosmatd}
\end{subfigure}
\caption{Cosine similarity for different situations (a) without permutation (b) with permutation \eqref{eq:neuperm} and (c) with fine-tuning and permutation \eqref{eq:neuperm} (d) is $P(\pi_l)$ for both (b) and (c). For visibility purposes, (b), (c), and (d) are clipped (first $100$ elements).}
\label{fig:cosmat}
\end{figure*}

With the previous approaches, we have observed that it is, in general, difficult to learn some input $\boldsymbol{y}_{l-1}$ such that the output $\boldsymbol{y}_{l}$ provides us the ranking of the neurons for $l$, as this approach is very sensitive to any minor perturbation introduced in the model. We have observed, though, that it is possible to uniquely recognize each neuron independently, learning a specific $\boldsymbol{y}_{l-1}$ which activates the target neuron. However, this solution consumes a lot of memory and computation resources. We have seen that two neurons, despite having the same output in some subspace of the trained domain, are in general very different. In particular, we can expect that $S_C(\boldsymbol{w}_{l,i,-},\boldsymbol{w}_{l,j,-}) < 1 \forall j\neq i$.\\
Let us study this phenomenon practically. Fig.~\ref{fig:cosmata} shows the correlation between the neuron's weights $\boldsymbol{w}_{l}$: since it is essentially a diagonal matrix, after applying some unknown permutation $\pi_l$ as in Fig.~\ref{fig:cosmatb}, we can easily recover the original positions building the Permutation matrix
\begin{equation}
    \label{eq:permobt}
    (P_{\pi_l})_{i,j} = \left\{
    \begin{array}{ll}
    1 & j = \text{argmax}_k\left[S_C(\boldsymbol{w}_{l,i,-}, \boldsymbol{w}_{l,k,-}^{\pi_l})\right]\\
    0 & \text{otherwise}.
    \end{array} \right .
\end{equation}
The question is here whether, even after applying some perturbation to the parameters, we are still able to recover the permutation $\pi$. As such, let us define $\tilde{\boldsymbol{w}}_{l,i,-}$ the set of parameters of the $i$ neuron in the $l$-th layer undergoing some perturbation. We can assume that any perturbation we want to introduce does not significantly change the performance $\mathcal{L}_{\Xi}$ of the trained model. As such, let us evaluate the cosine similarity between $\boldsymbol{w}_{l,i,-}$ and $\tilde{\boldsymbol{w}}_{l,i,-}$: we expect that when this measure drops, the performance of the model will drop as well. Two neurons, despite having the same output in the trained domain, are in general different: we can expect that 
\begin{equation}
\label{eq:condition}
S_C(\boldsymbol{w}_{l,i,-},\tilde{\boldsymbol{w}}_{l,i,-}) > S_C(\boldsymbol{w}_{l,i,-},\tilde{\boldsymbol{w}}_{l,j,-})\forall j\neq i.
\end{equation}
According to \eqref{eq:condition}, it is possible to detect where the $i$-th neuron has been displaced, thus, recovering the original ordering. This condition obeys some theoretical warranties.
Let us compare the set parameters $\boldsymbol{w}_{l,i}$ to the same, where we apply a perturbation, which results in $\tilde{\boldsymbol{w}}_{l,i} = \boldsymbol{w}_{l,i} + \hat{\boldsymbol{w}}_{l,i}$. According to the Cauchy-Schwarz inequality, the only possible solution is that $\hat{\boldsymbol{w}}_{l,i}$ is a scalar multiple of $\boldsymbol{w}_{l,i}$. \\
Let us investigate the case in which we perform fine-tuning on the parameters: we record a slight improvement in the performance with $\Theta=2\%$, and we observe that the permutation matrix (Fig.~\ref{fig:cosmatd}) we obtain from the cosine similarities (Fig.~\ref{fig:cosmatc}) is the same as the one recovered before, making out re-synchronization success rate to $100\%$.
The details of our method are presented in Alg.~\ref{alg:np} and Fig.~\ref{fig:flo}.

\subsection{Integrity loss}
\label{sec:counter-counter-attack}
We will analyze here the special case when $\boldsymbol{\hat{w}}_{l,i} = k\cdot \boldsymbol{w}_{l,i}$.  \\
 Let us assume the input of the $l$-th layer follows a Gaussian distribution, with mean $\boldsymbol{\mu}_l$ and covariance matrix $\Sigma_l$. We know that the post-synaptic potential still follows a Gaussian distribution $\mathcal{N}(\mu_z,\sigma_z^2)$. Given that $\boldsymbol{\hat{w}}_{l,i}$ will produce as output $\tilde{z}$, we can write the KL-divergence between the outputs generated from the original and from the perturbed neuron
\begin{equation}
    \label{eq:KLdivergence}
    D_{\text{KL}}(z || \tilde{z}) = \log (1+k) + \frac{\sigma_z^2+k^2\mu_z^2}{2(1+k)^2 \sigma_z^2} - \frac{1}{2}.
\end{equation}
Under the assumption of having an activation such that $|\varphi(x)'| \leq 1 \forall x \in \mathbb{R}$, we know that the above divergence upper-bounds $D_{\text{KL}}(y || \tilde{y})$. Specifically, for ReLU activations, under the assumption of $\mu_z = 0$, the KL-divergence is
\begin{equation}
    \label{eq:klrelu}
    D_{\text{KL}}(y || \tilde{y}) =\frac{2(k+1)^2\log(k+1) - k(k+2)}{(k+1)^2},
\end{equation}
which is dependent on $k$ only. Despite having maximum similarity (except for the degenerate case $k=-1$), the KL divergence of the output is non-zero $\forall k\neq 0$, which means the behavior of the model is modified. 

%% file: sections/4_experiments.tex
\section{Experimental results}
\label{sec:experiments}
\begin{table*}
\centering
\caption{Robustness to Gaussian noise addition.} \label{tab:addgauss}
\resizebox{0.95\textwidth}{!}{
\begin{tabular}{ c c c c c c c c c c c}
 \toprule
    \multirow{2}{*}{$\boldsymbol{\Omega}$}&&\multicolumn{2}{c}{\bf CIFAR10}&\multicolumn{4}{c}{\bf ImageNet}&\bf Cityscapes& \bf COCO & \bf UVG \\
    & & VGG16 & ResNet18 & ResNet50 & ResNet101 & ViT-b-32 & MobileNetV3 &DeepLab-v3&YOLO-v5n & DVC\\
\midrule
 \multirow{2}{*}{$0$} & $\Psi (\uparrow)$ & 100\small{$\pm0$} & 100\small{$\pm0$} & 100 & 100 & 100 & 100 & 100 & 100 & 100 \\
 & \it metric$(\downarrow)$ &  \it 9.96$^\dag$\small{$\pm0.19$} & \it 7.03$^\dag$\small{$\pm0.28$} & \it 23.85$^\dag$ & \it 22.63$^\dag$ & \it 24.07$^\dag$  & \it 25.95$^\dag$ & \it 32.26$^\ddag$ & \it 48.70$^\star$ & \it 0.177$^\bullet$\\
\midrule
\multirow{2}{*}{$1$} & $\Psi (\uparrow)$ & 100\small{$\pm0$} & 100\small{$\pm0$} & 100 & 100 & 100 & 100 & 100 & 75.69 & 100\\
 & \it metric$(\downarrow)$ & \it 10.25$^\dag$\small{$\pm0.17$} & \it 7.30$^\dag$\small{$\pm0.08$} & \it 24.73$^\dag$    & \it 23.17$^\dag$ & \it 24.08$^\dag$ & \it 40.44$^\dag$ & \it 32.16$^\ddag$ & \it 79$^\star$ & \it 0.177$^\bullet$\\
\midrule
  \multirow{2}{*}{$2$} & $\Psi (\uparrow)$ & 100\small{$\pm0$} & 100\small{$\pm0$} & 99.56 & 99.26 & 99.64 & 100 & 100 & 57.65 & 100\\
 & \it metric$(\downarrow)$& \it 11.82$^\dag$\small{$\pm0.17$} & \it 9.13$^\dag$\small{$\pm0.59$} & \it 27.68$^\dag$  & \it 25.08$^\dag$ & \it 24.77$^\dag$ & \it 70.50$^\dag$ & \it 32.75$^\ddag$ & \it 82.10$^\star$ & \it 0.177$^\bullet$ \\
\midrule
  \multirow{2}{*}{$7$} & $\Psi (\uparrow)$ & 99.88\small{$\pm0.12$} & 99.90\small{$\pm0.10$} & 57.28  & 39.45 & 99.64 & 85.31 & 41.02 & 8.24 & 100\\
 & \it metric$(\downarrow)$& \it 41.27$^\dag$\small{$\pm2.11$}  & \it 99.55$^\dag$\small{$\pm6.30$} & \it 66.97$^\dag$ & \it 60.49$^\dag$ & \it 60.39$^\dag$ &\it 98.91$^\dag$ & \it 38.30$^\ddag$ & \it 99.62$^\star$ & \it 0.180$^\bullet$ \\
  \midrule
\multirow{2}{*}{$10$} & $\Psi (\uparrow)$ & 93.50\small{$\pm0.98$} & 94.9\small{$\pm0.80$} & 12.93 & 12.74 & 99.22 & 43.05 & 12.89 & 5.49 & 100\\
 & \it metric$(\downarrow)$ & \it 56.62$^\dag$\small{$\pm2.31$} & \it 75.18$^\dag$\small{$\pm5.14$} & \it 92.65$^\dag$ & \it 83.04$^\dag$ &\it 83.99$^\dag$ &\it 99.41$^\dag$ & \it 39.74$^\ddag$ & \it 99.23$^\star$ & \it 0.182$^\bullet$\\
 
  \bottomrule
 \end{tabular}
}
\end{table*}

\begin{table*}
\centering
\caption{Robustness to fine-tuning.} \label{tab:finetuning}
\resizebox{0.95\textwidth}{!}{
\begin{tabular}{ c c c c c c c c c c c}
 \toprule
    \multirow{2}{*}{$\boldsymbol{\Theta}$}&&\multicolumn{2}{c}{\bf CIFAR10}&\multicolumn{4}{c}{\bf ImageNet}&\bf Cityscapes& \bf COCO & \bf UVG \\
    & & VGG16 & ResNet18 & ResNet50 & ResNet101 & ViT-b-32 & MobileNetV3 &DeepLab-v3&YOLO-v5n & DVC\\
 \midrule
 \multirow{2}{*}{$2$} & $\Psi (\uparrow)$ & 100\small{$\pm0$} & 100\small{$\pm0$} & 100 & 100 & 100 & 100 & 100 & 100 & 100\\
  & \it metric$(\downarrow)$  & \it 9.97$^\dag$\small{$\pm0.20$} & \it 6.96$^\dag$\small{$\pm0.11$} & \it 23.35$^\dag$ & \it 22.54$^\dag$ & \it 24.08$^\dag$ & \it 26.60$^\dag$  &\it 34.85$^\ddag$ & \it 47.40$^\star$ & \it 0.184$^\bullet$ \\

\midrule
  \multirow{2}{*}{$6$} & $\Psi (\uparrow)$ & 100\small{$\pm0$} & 100\small{$\pm0$} & 100 & 100 & 100 & 100 & 100 & 100 & 100\\
 & \it metric$(\downarrow)$ & \it 9.96$^\dag$\small{$\pm0.19$} & \it 6.98$^\dag$\small{$\pm0.15$}  & \it 23.23$^\dag$ & \it 22.57$^\dag$ & \it 24.08$^\dag$ & \it 26.41$^\dag$ &\it 31.58$^\ddag$ & \it 46.20$^\star$ & \it 0.179$^\bullet$ \\
\midrule
\multirow{2}{*}{$8$} & $\Psi (\uparrow)$ & 100\small{$\pm0$} & 100\small{$\pm0$} & 100 & 100 & 100 & 100 & 100 & 100 & 100\\
 & \it metric$(\downarrow)$ & \it 9.88$^\dag$\small{$\pm0.24$} & \it 7.01$^\dag$\small{$\pm0.19$}  & \it 23.21$^\dag$ & \it 22.54$^\dag$ & \it 24.07$^\dag$ & \it 26.37$^\dag$ &\it 30.35$^\ddag$ & \it 46.40$^\star$ & \it 0.179$^\bullet$ \\
\midrule
 \multirow{2}{*}{$10$} & $\Psi (\uparrow)$ & 100\small{$\pm0$} & 100\small{$\pm0$} & 100 & 100 & 100 & 100 & 100 & 100 & 100\\
 & \it metric$(\downarrow)$ & \it 9.89$^\dag$\small{$\pm0.16$} & \it 6.94$^\dag$\small{$\pm0.13$}   & \it 23.13$^\dag$ & \it 22.49$^\dag$ & \it 24.07$^\dag$ & \it 26.31$^\dag$ &\it 29.64$^\ddag$ & \it 46.00$^\star$ & \it 0.178$^\bullet$ \\

  \bottomrule
 \end{tabular}
}
\end{table*}

\paragraph{Datasets} We will test our proposed approach on five datasets: CIFAR-10~\cite{krizhevsky2009learning} and ImageNet-1k~\cite{ILSVRC15} for image classification (metric is here top-1 classification error denoted by $\dag$), CityScapes~\cite{cordts2016cityscapes} for image segmentation (metric is here the complementary mean IoU $\ddag$), COCO~\cite{lin2014microsoft} for object detection (metric is here the complementary of mAP50~$\star$) and UVG~\cite{mercat2020uvg} (metric is here the mean rate-distortion (bpp) $\bullet$ for a given image quality, MS-SSIM $=0.97$).

\paragraph{Implementation details} We evaluate our approach on many different state-of-the-art architectures: VGG-16~\cite{simonyan2014very}, ResNet18~\cite{he2016deep}, ResNet50~\cite{he2016deep}, ResNet101~\cite{he2016deep}, ViT-b-32~\cite{dosovitskiy2020image}, MobileNetV3~\cite{howard2019searching}, DeepLabV3~\cite{chen2018encoder}, YOLO-v5n~\cite{glenn_jocher_2022_7002879} and DVC~\cite{lu2019dvc}. We will test the robustness of our approach using the re-synchronization success rate $\Psi$ after applying random permutation to the penultimate layer and the four perturbations. For all of the aforementioned experiments, we have used all the traditional setups described in the respective original papers. We further notice that for the models trained on CIFAR-10, we have run 10 seeds, and average results are reported.\footnote{the source code is available at \url{https://github.com/carldesousatrias/FindtheLady}}

\paragraph{Robustness against Gaussian noise} We evaluate our methods against Gaussian noise addition with $\Omega \in [1,10]$. The error starts increasing while $\Psi$ remains very close to $100\%$. For instance, $\Omega$ valuing 6, $\Psi$ is still equal to $100\%$ while the error has more than doubled. Table~\ref{tab:addgauss} reports the results for all the architectures and the datasets. In particular, we observe that consistently for all the architectures except YOLO, when the error starts increasing, $\Psi$ remains very close to $100\%$. But for YOLO, we observe the error has more than doubled while we only failed to recover a fifth of the original order.

\begin{table*}
\centering
\caption{Robustness to quantization.} \label{tab:quantiz}
\resizebox{0.95\textwidth}{!}{
\begin{tabular}{ c c c c c c c c c c c}
 \toprule
    \multirow{2}{*}{$\boldsymbol{B}$}&&\multicolumn{2}{c}{\bf CIFAR10}&\multicolumn{4}{c}{\bf ImageNet}&\bf Cityscapes& \bf COCO & \bf UVG \\
    & & VGG16 & ResNet18 & ResNet50 & ResNet101 & ViT-b-32 & MobileNetV3 &DeepLab-v3&YOLO-v5n & DVC\\
 \midrule

 \multirow{2}{*}{$16$} & $\Psi (\uparrow)$ & 100\small{$\pm0$} & 100\small{$\pm0$} & 100 & 100 & 100 & 100&  100 & 100 & 100\\
 & \it metric$(\downarrow)$ & \it 9.96$^\dag$\small{$\pm0.19$} & \it 7.03$^\dag$\small{$\pm0.28$} &  \it 23.85$^\dag$ & \it 22.63$^\dag$ & \it 24.08$^\dag$& \it 25.96$^\dag$  & \it 32.26$^\ddag$ & \it 48.70$^\star$ & \it 0.177$^\bullet$\\
 \midrule
\multirow{2}{*}{$8$} & $\Psi (\uparrow)$ & 100\small{$\pm0$} & 100\small{$\pm0$} & 100 & 100 & 100 & 100 & 100 & 100 & 98.44\\
 & \it metric$(\downarrow)$& \it 9.97$^\dag$\small{$\pm0.20$} & \it 7.05$^\dag$\small{$\pm0.27$} & \it 23.91$^\dag$ & \it 22.70$^\dag$ & \it 24.10$^\dag$ & \it 26.00$^\dag$ & \it  31.49$^\ddag$ & \it 48.90$^\star$ & \it 0.338$^\bullet$ \\
 \midrule
  \multirow{2}{*}{$6$} & $\Psi (\uparrow)$ & 100\small{$\pm0$} & 100\small{$\pm0$} & 100 & 100 & 100 & 100  & 100 & 100 & 98.44\\
 & \it metric$(\downarrow)$& \it 9.98$^\dag$\small{$\pm0.17$} & \it 7.14$^\dag$\small{$\pm0.27$} & \it 26.99$^\dag$ & \it 25.12$^\dag$ & \it 29.55$^\dag$ & \it 42.91$^\dag$ & \it 32.26$^\ddag$ & \it 89.10$^\star$ & \it 0.823$^\bullet$\\
 \midrule
 \multirow{2}{*}{$4$} & $\Psi (\uparrow)$ & 100\small{$\pm0$} & 100\small{$\pm0$} & 100 & 100 & 100 & 100 & 100 & 85.49 & 98.44\\
 & \it metric$(\downarrow)$& \it 10.77$^\dag$\small{$\pm0.28$} & \it 7.76$^\dag$\small{$\pm0.26$} & \it 99.91$^\dag$ & \it 99.99$^\dag$  & \it 96.81$^\dag$ & \it 99.91$^\dag$ & \it 47.28$^\ddag$ & \it 100$^\star$ & \it $\infty^\bullet$\\
 \midrule
 \multirow{2}{*}{$2$} &  $\Psi (\uparrow)$ & 100\small{$\pm0$} & 100\small{$\pm0$} & 31.54 & 9.91 & 100 & 100 & 100 & 56.47 & 98.44\\
 & \it metric$(\downarrow)$& \it 87.73$^\dag$\small{$\pm5.56$} & \it 88.20$^\dag$\small{$\pm2.77$} & \it 99.9$^\dag$ & \it 99.9$^\dag$ & \it 99.81$^\dag$ & \it 99.9$^\dag$ & \it 96.63$^\ddag$ & \it 100$^\star$ & \it $\infty^\bullet$\\

  \bottomrule
 \end{tabular}
}
\end{table*}
\begin{table*}
\centering
\caption{Robustness to magnitude pruning.} \label{tab:pruning}
\resizebox{0.95\textwidth}{!}{
\begin{tabular}{ c c c c c c c c c c c}
 \toprule
    \multirow{2}{*}{$\boldsymbol{T}$}&&\multicolumn{2}{c}{\bf CIFAR10}&\multicolumn{4}{c}{\bf ImageNet}&\bf Cityscapes& \bf COCO & \bf UVG \\
    & & VGG16 & ResNet18 & ResNet50 & ResNet101 & ViT-b-32 & MobileNetV3 &DeepLab-v3&YOLO-v5n & DVC\\
 \midrule
   \multirow{2}{*}{$0$} & $\Psi (\uparrow)$ & 100\small{$\pm0$} & 100\small{$\pm0$} & 100 & 100 & 100 & 100 & 100 & 100 & 100\\
 & \it metric$(\downarrow)$ & \it 9.96$^\dag$\small{$\pm0.19$} & \it 7.03$^\dag$\small{$\pm0.28$}& \it 23.85$^\dag$ & \it 22.63$^\dag$ & \it 24.07$^\dag$  & \it 25.95$^\dag$ & \it 32.26$^\ddag$ & \it 48.70$^\star$ & \it 0.177$^\bullet$\\
 \midrule
 \multirow{2}{*}{$0.91$} & $\Psi (\uparrow)$ & 100\small{$\pm0$} & 100\small{$\pm0$} & 98.87 & 98.34 & 100 & 100 & 100 & 65.88 & 100\\
 & \it metric$(\downarrow)$ & \it 10.87$^\dag$\small{$\pm0.22$} & \it 9.67$^\dag$\small{$\pm0.95$} & \it 26.30$^\dag$ & \it 23.94$^\dag$ & \it 25.06$^\dag$ & \it 40.00$^\dag$ & \it 33.11$^\ddag$ & \it 50$^\star$ & \it 0.188$^\bullet$\\
 \midrule
\multirow{2}{*}{$0.95$} & $\Psi (\uparrow)$ & 100\small{$\pm0$} & 100\small{$\pm0$} & 97.61 & 97.12 & 100 & 100 & 100 & 51.76 & 100\\
 & \it metric$(\downarrow)$& \it 12.11$^\dag$\small{$\pm0.45$} & \it 12.88$^\dag$\small{$\pm1.90$} & \it 28.35$^\dag$ & \it 24.58$^\dag$ &\it 25.60$^\dag$ & \it 48.73$^\dag$ & \it 33.75$^\ddag$ & \it 51.10$^\star$ & \it 0.191$^\bullet$ \\
 \midrule
  \multirow{2}{*}{$0.98$} & $\Psi (\uparrow)$ & 100\small{$\pm0$} & 100\small{$\pm0$} & 93.12 & 91.94 & 99.83 & 97.91 & 99.61 & 36.47 & 100\\
 & \it metric$(\downarrow)$& \it 17.53$^\dag$\small{$\pm1.28$}   & \it 21.71$^\dag$\small{$\pm5.39$} & \it 30.88$^\dag$ & \it 25.65$^\dag$ & \it 26.03$^\dag$ & \it 63.48$^\dag$ & \it 34.47$^\ddag$ & \it 52.60$^\star$ & \it 0.198$^\bullet$\\
  \midrule
\multirow{2}{*}{$0.99$} & $\Psi (\uparrow)$ & 99.90\small{$\pm0.13$} & 99.63\small{$\pm0.31$} & 81.10 & 78.81 & 95.37 & 86.46 & 90.24 & 25.88 & 100\\
 & \it metric$(\downarrow)$& \it 26.71$^\dag$\small{$\pm2.84$}  & \it 28.16$^\dag$\small{$\pm7.35$}  & \it 32.62$^\dag$ & \it 25.98$^\dag$ & \it 26.63$^\dag$ & 76.22$^\dag$  & \it 36.43$^\ddag$ & \it 41.40$^\star$ & \it 0.219$^\bullet$\\

  \bottomrule
 \end{tabular}
}
\end{table*}

\paragraph{Robustness against fine-tuning} We here evaluate our method against perturbations produced by simply fine-tuning the model, adding more training complexity. Table~\ref{tab:finetuning} presents the results for all the architectures. In particular, we observe that consistently for all the architectures, $\Psi$ remains equal to $100\%$. Despite, different experimental setups, the error on YOLO-v5n always increases: so we decided to not extend its test. 

\paragraph{Robustness against quantization} We evaluate our methods against quantization. In particular, we will evaluate the performance with $B\in [2; 16]$. Specifically, the error starts increasing around 3 bits while $\Psi$ remains very close to $100\%$. A plot is provided as an Appendix. 
Table~\ref{tab:quantiz} presents the results for all architectures. Remarkably, for most of the architectures, including YOLO and ViT, $\Psi$ remains close to $100\%$ despite the error being extremely high.

\paragraph{Robustness against magnitude pruning} We evaluate our methods against magnitude pruning $T \in [90; 99]\%$ of the aimed layer. The error starts increasing while $\Psi$ remains very close to $100\%$. Table~\ref{tab:pruning} shows good robustness for most of the architectures. For the ResNet and YOLO structures, $\Psi$ decrease before having a huge increase of the error, but, even if the aimed layer is fully pruned the error rate remains below $50\%$ and $70\%$ respectively: this is due to the residual connections. For both, we did additional experiments applying global pruning to the model and the performances are similar to the other architectures.

\begin{figure}[t]
\centering
\includegraphics[width=0.85\columnwidth, trim=10 0 10 18, clip]{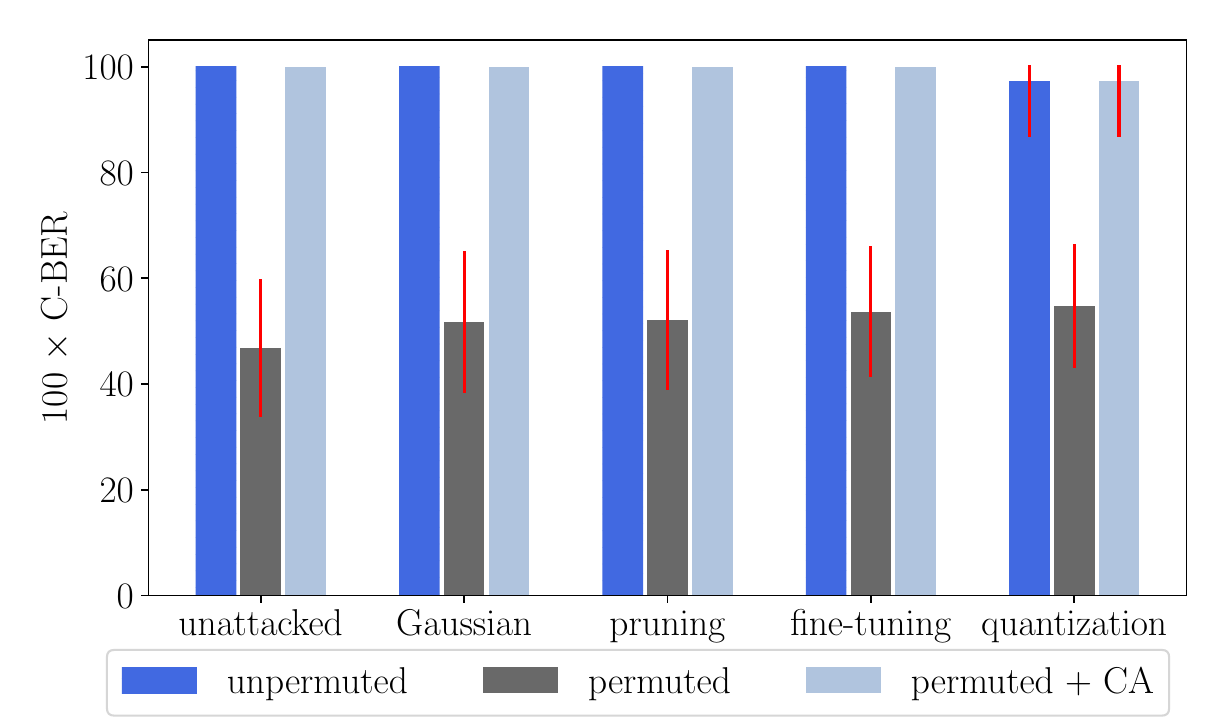}
\caption{Robustness evaluation of the~\cite{uchida2017embedding}'s watermarking method against the 4 attacks. }
\label{fig:water}
\end{figure}

\paragraph{Application to white-box watermarking} Watermarking of neural networks is increasingly considered an important problem with many practical applications (the challenge of watermarking ChatGPT or assessing the integrity of unmanned vehicles). Currently, the white-box watermarking literature fails to be robust against permutation attacks.
Fig.~\ref{fig:water} shows the correlation (evaluated as Pearson correlation coefficient) of a white-box watermark when employing a state-of-the-art approach~\cite{uchida2017embedding}. Uchida et al.'s approach is considered one of the first white-box watermarking methods, where a regularization term is added to the cost function to change the distribution of one pre-selected layer in the model. It projects the parameter of the watermarked layer on a space a binary watermark. The order of neurons is mandatory to recover the original binary mark. We observe that permuted neurons, although not impacting the performance of the model, destroy the correlation. Applying our approach as a counter-attack (CA), we observe that we successfully retrieve the watermark and preserve the robustness, more applicative results are presented in~\cite{de2023hitchhiker}.

\begin{figure}[t]
\centering
\begin{subfigure}[h]{0.45\columnwidth}
    \includegraphics[width=\textwidth]{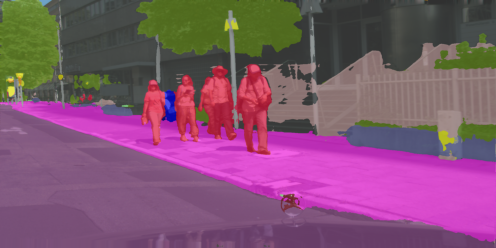}
    \caption{Original output.}
\end{subfigure}
\begin{subfigure}[h]{0.45\columnwidth}
    \includegraphics[width=\textwidth]{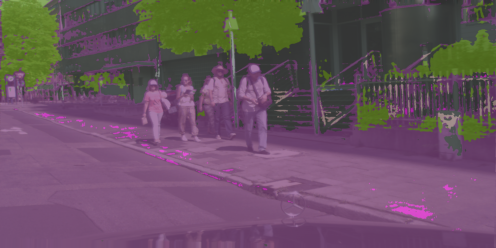}
    \caption{Altered output.}
\end{subfigure}
\caption{Misdetection of the pedestrian induced by the scalar product modification of the weights.}
\label{fig:citys}
\end{figure}

\paragraph{Integrity loss} Let us here consider a counter-attack for our algorithm, on a real application: pedestrians are not detected anymore while the cosine similarity remains still equal to one (the effect in Fig.~\ref{fig:citys}). To protect our method against this issue, we simply need to add a $\ell_2$-norm verification between $\boldsymbol{w}_{l,i}$ and $\tilde{\boldsymbol{w}}_{l,i}$: any modification to the norm can, in this way, detected and corrected.

%% file: sections/5_conclusion.tex
\section{Conclusion}
\label{sec:conclusion}
In this paper, we have defined and investigated one uprising question for deep learning models: is it possible to recognize parameters in a neuron after some perturbations? Is it possible to recover an original ordering for the neurons after random permutations and some perturbations? 

\noindent We have explored the realm of neuron similarity, observing the parameters and outputs of different layers. We have investigated many ways to do so, observing and assessing their failure reasons. Finally, we advance a method that leverages the cosine similarity between the original layer and its permuted, perturbed version. We empirically assessed the robustness of this approach against several perturbations, for a variety of architectures and datasets. 

\noindent This work has a direct impact on watermarking, where it serves as a generic counter-attack tool against parameter permutation, and has an indirect impact in various other AI domains, like pruning: as a result, neurons having perfectly correlated outputs typically have orthogonal kernels.

%% file: supplementary.tex
\appendix
\onecolumn 

\section{Table of notation}
In Table~\ref{tab:notation}, we present the table of notation used in the main document.
\begin{table}[h]
    \centering
    \caption{Table of notation.}
    \label{tab:notation}
    %\resizebox{0.95\columnwidth}{!}{
        \begin{tabular}{ c l }
            \toprule
            \bf Symbol & \bf Definition \\
            \midrule
                $\boldsymbol{y}_{l}$ & output of the $l$-th layer\\
             %\midrule
                $\boldsymbol{w}_l$ & weights of the $l$-th layer of the model \\
               %\midrule
               $\left<\cdot\right>$ & the inner product \\
               %\midrule
               $(\cdot)^T$ & the transpose operator \\
               %\midrule
                $\varphi(\cdot)$ & the activation function \\
               %\midrule
                $\boldsymbol{w}_{l,c,-}$ & all elements of the $l$-th layer, for the $c$-th channel\\
              %\midrule
                $\boldsymbol{w}_{l,-,n}$  & all elements of the $l$-th layer, for the $n$-th neurons \\
               %\midrule
                \multirow{2}{*}{$P_{\pi_l}$} & permutation matrix (square matrix of size\\
                            &  number of neurons) \\
                %\midrule
                $D_{\text{KL}}$ &  the Kullback–Leibler divergence \\
                %\midrule
                $\Xi$ & the dataset the model is trained on \\
                %\midrule
                $S_C$ & cosine similarity matrix \\
                %\midrule
                $\Psi$ & re-synchronization success rate \\
                %\midrule
                $\Omega$ & the \textit{power} of the Gaussian noise \\
                %\midrule
                $\Theta$ & percentage of additional training \\
                %\midrule
                $\boldsymbol{B}$ & number of bits \\
                %\midrule
                $\boldsymbol{T}$ & fraction of masked weights \\
                %\midrule
            \bottomrule
        \end{tabular}
    %}
\end{table}

\section{Toy example of permutation}

\begin{figure}[h]
    \centering
    \begin{subfigure}[b]{0.2\textwidth}
        \centering
        \includegraphics[width=\textwidth]{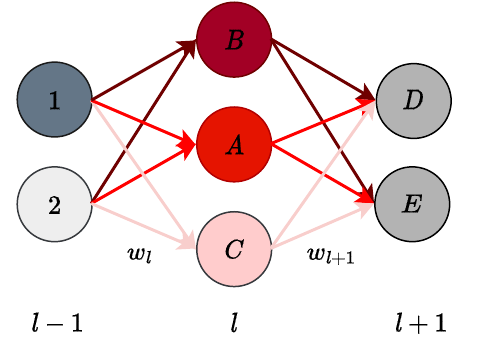}
        \caption{original state}
        \label{fig2:perma}
    \end{subfigure}
    \hfill
    \begin{subfigure}[b]{0.2\textwidth}
        \centering
        \includegraphics[width=\textwidth]{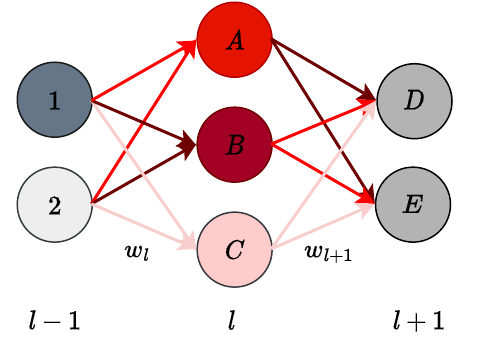}
        \caption{neuron permutation}
        \label{fig2:permb}
    \end{subfigure}
    \hfill
    \begin{subfigure}[b]{0.2\textwidth}
        \centering
        \includegraphics[width=\textwidth]{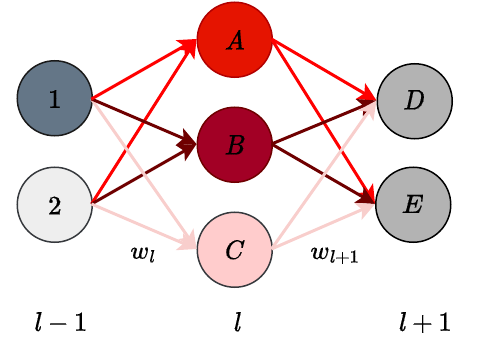}
        \caption{channel permutation}
        \label{fig2:permc}
    \end{subfigure}
    \caption{Illustration of the permutation process on a fully connected neural network of three layers}
    \label{fig2:permutation}
\end{figure}

In this section, we take a simple example to describe the permutation problem using a three-layer model, Fig.~\ref{fig2:perma}. It is a fully connected model without bias and we observe the final output of this model: 
$\boldsymbol{y}_{l+1}\in \mathbb{R}^{2 \times 1}$. The equation (1) become:
\begin{equation}
\label{eq2:base}
    \begin{pmatrix}
    y_{l+1,D}\\
    y_{l+1,E}
    \end{pmatrix}
    = \phi  \left [
    \begin{pmatrix}
        \textcolor{purple}{\boldsymbol{w}_{l+1,B,D}} & \textcolor{red}{\boldsymbol{w}_{l+1,A,D}} & \textcolor{magenta}{\boldsymbol{w}_{l+1,C,D}} \\
        \textcolor{purple}{\boldsymbol{w}_{l+1,B,E}} & \textcolor{red}{\boldsymbol{w}_{l+1,A,E}} & \textcolor{magenta}{\boldsymbol{w}_{l+1,C,E}}
    \end{pmatrix} \cdot
    \begin{pmatrix}
    \textcolor{purple}{y_{l,B}}\\
    \textcolor{red}{y_{l,A}}\\
    \textcolor{magenta}{y_{l,C}}
    \end{pmatrix} \right]
\end{equation}
where $\boldsymbol{y}_{l}\in \mathbb{R}^{3 \times 1}$ is the input of the $l+1$-th layer, and $\boldsymbol{w}_{l+1}\in \mathbb{R}^{2\times 3}$ are the weights for the $l+1$-th layer. We can use again equation (1) to include the expression of the $l$-th layer in equation (\ref{eq2:base}):
\begin{equation}
    \begin{pmatrix}
        y_{l+1,D}\\
        y_{l+1,E}
    \end{pmatrix} 
    = \phi  \left \{
    \begin{pmatrix}
        \textcolor{purple}{\boldsymbol{w}_{l+1,B,D}} & \textcolor{red}{\boldsymbol{w}_{l+1,A,D}} & \textcolor{magenta}{\boldsymbol{w}_{l+1,C,D}} \\
        \textcolor{purple}{\boldsymbol{w}_{l+1,B,E}} & \textcolor{red}{\boldsymbol{w}_{l+1,A,E}} & \textcolor{magenta}{\boldsymbol{w}_{l+1,C,E}}
    \end{pmatrix} \cdot  ~\phi  \left[
    \begin{pmatrix}
        \color{purple} \boldsymbol{w}_{l,1,B} & \color{purple} \boldsymbol{w}_{l,2,B} \\
        \color{red} \boldsymbol{w}_{l,1,A} & \color{red} \boldsymbol{w}_{l,2,A} \\
        \color{magenta} \boldsymbol{w}_{l,1,C} & \color{magenta} \boldsymbol{w}_{l,2,C} 
    \end{pmatrix}
    \begin{pmatrix}
        y_{l-1,1}\\
        \textcolor{gray}{y_{l-1,2}}
    \end{pmatrix} 
     \right] \right\}
    \label{eq2:base2}
\end{equation}
where $\boldsymbol{y}_{l}\in \mathbb{R}^{3 \times 1}$ is the input of the $l$-th layer $\boldsymbol{w}_{l}\in \mathbb{R}^{3\times 2}$ are the weights for the $l$-th layer.
Let us consider the case a permutation between neuron $A$ and neuron $B$ ($\pi_l$) is applied on the neurons of the $l$-th layer; the permutation matrix is 
\begin{equation}
     P_{\pi_l}=\left(\begin{matrix}
0 & 1 &  0\\
1 & 0 & 0\\
0 & 0 & 1\\
\end{matrix}\right)
\label{eq2:perm}
\end{equation}
The neurons are permuted, and the ordering for the input channels $\boldsymbol{y}_{l}$ remains intact (Fig.~\ref{fig2:permb}). Hence, the permuted output for the $l+1$-th layer will be:
\begin{equation}
    \begin{pmatrix}
        y_{l+1,D}^{\pi_l}\\
        y_{l+1,E}^{\pi_l}
    \end{pmatrix} 
    = \phi   \left \{
    \begin{pmatrix}
        \textcolor{purple}{\boldsymbol{w}_{l+1,B,D}} & \textcolor{red}{\boldsymbol{w}_{l+1,A,D}} & \textcolor{magenta}{\boldsymbol{w}_{l+1,C,D}} \\
        \textcolor{purple}{\boldsymbol{w}_{l+1,B,E}} & \textcolor{red}{\boldsymbol{w}_{l+1,A,E}} & \textcolor{magenta}{\boldsymbol{w}_{l+1,C,E}}
    \end{pmatrix} \cdot  ~\phi  \left[
    \begin{pmatrix}
        \color{red} \boldsymbol{w}_{l,1,A} & \color{red} \boldsymbol{w}_{l,2,A} \\
        \color{purple} \boldsymbol{w}_{l,1,B} & \color{purple} \boldsymbol{w}_{l,2,B} \\
        \color{magenta} \boldsymbol{w}_{l,1,C} & \color{magenta} \boldsymbol{w}_{l,2,C} 
    \end{pmatrix}
    \begin{pmatrix}
        y_{l-1,1}\\
        \textcolor{gray}{y_{l-1,2}}
    \end{pmatrix} 
     \right] \right\}
    \label{eq2:neuperm}
\end{equation}
After having simply applied $\pi_l$ at layer $l$, however, the output of the model is likely to be altered, as the propagated $\boldsymbol{y}_{l}^{\pi_l}\neq \boldsymbol{y}_{l}$, which is processed as input by the next layer. We know, however, that $\boldsymbol{y}_{l}^{\pi_l}$ is a permutation of $\boldsymbol{y}_{l}$; hence, to maintain the output of the full model unaltered, we need to also permute the weights in layer $l+1$ as
\begin{equation}
    \begin{pmatrix}
        y_{l+1,D}^{\pi_l}\\
        y_{l+1,E}^{\pi_l}
    \end{pmatrix} 
    = \phi  \left \{
    \begin{pmatrix}
        \textcolor{red}{\boldsymbol{w}_{l+1,A,D}} & \textcolor{purple}{\boldsymbol{w}_{l+1,B,D}} & \textcolor{magenta}{\boldsymbol{w}_{l+1,C,D}} \\
        \textcolor{red}{\boldsymbol{w}_{l+1,A,E}} & \textcolor{purple}{\boldsymbol{w}_{l+1,B,E}} & \textcolor{magenta}{\boldsymbol{w}_{l+1,C,E}}
    \end{pmatrix}  \cdot ~\phi  \left[
    \begin{pmatrix}
        \color{red}\boldsymbol{w}_{l,1,A} & \color{red}\boldsymbol{w}_{l,2,A} \\
        \color{purple} \boldsymbol{w}_{l,1,B} & \color{purple} \boldsymbol{w}_{l,2,B} \\
        \color{magenta} \boldsymbol{w}_{l,1,C} & \color{magenta}  \boldsymbol{w}_{l,2,C} 
    \end{pmatrix}
    \begin{pmatrix}
        y_{l-1,1}\\
        \textcolor{gray}{y_{l-1,2}}
    \end{pmatrix} 
     \right] \right\}
    \label{eq2:chanperm}
\end{equation}
In this way, the permuted outputs in the $l$-th layer will be correctly weighted in the next layer, and the neural network output will be unchanged (Fig.~\ref{fig2:permc}).

%\newpage
\section{Additional figures for the experimental section}
\begin{figure}[!h]
\centering
\begin{subfigure}{0.49\textwidth}%[h]
\includegraphics[width=\textwidth]{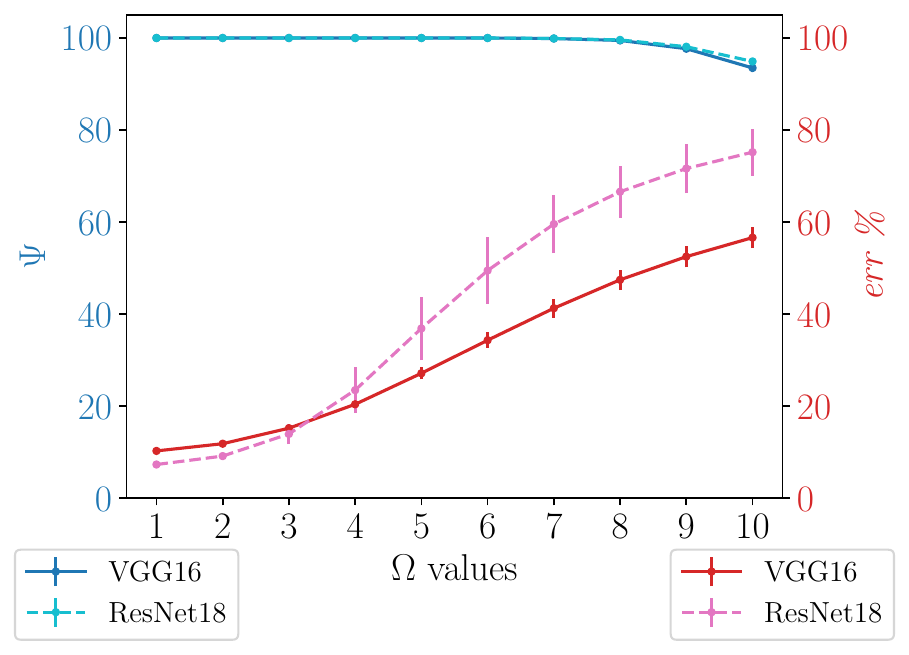}
\caption{Robustness against Gaussian noise addition. $\Psi$ is on the left axis in blue and the \textit{err \%} on the right axis in red.}
\label{fig:addgauss}
\end{subfigure}
~
\begin{subfigure}{0.49\textwidth}
\includegraphics[width=\textwidth]{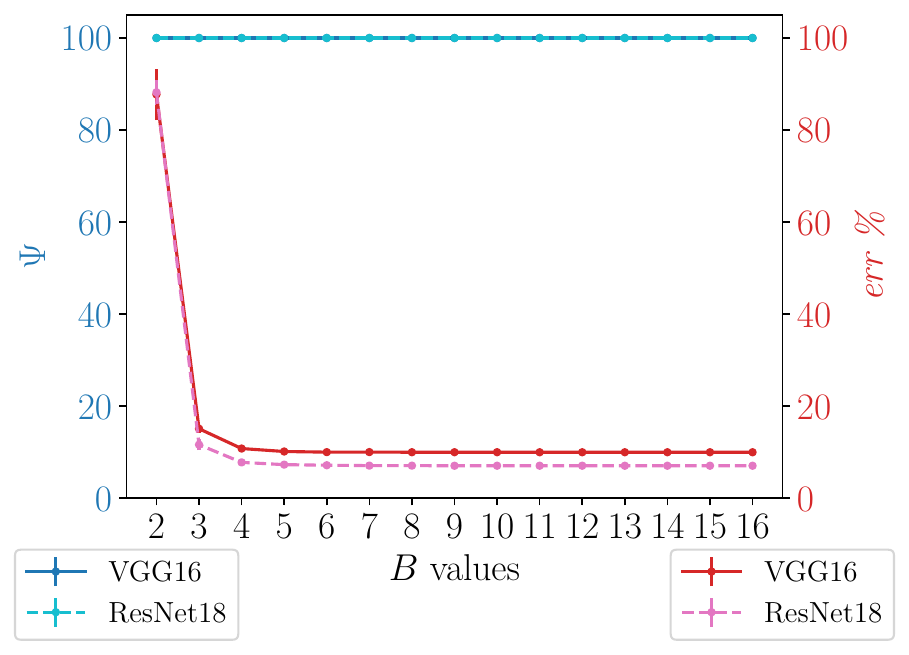}
\caption{Robustness against quantization. $\Psi$ is on the left axis in blue and the \textit{err \%} on the right axis in red.}
\label{fig:quantiz}
\end{subfigure}

\begin{subfigure}{0.49\textwidth}
\includegraphics[width=\textwidth]{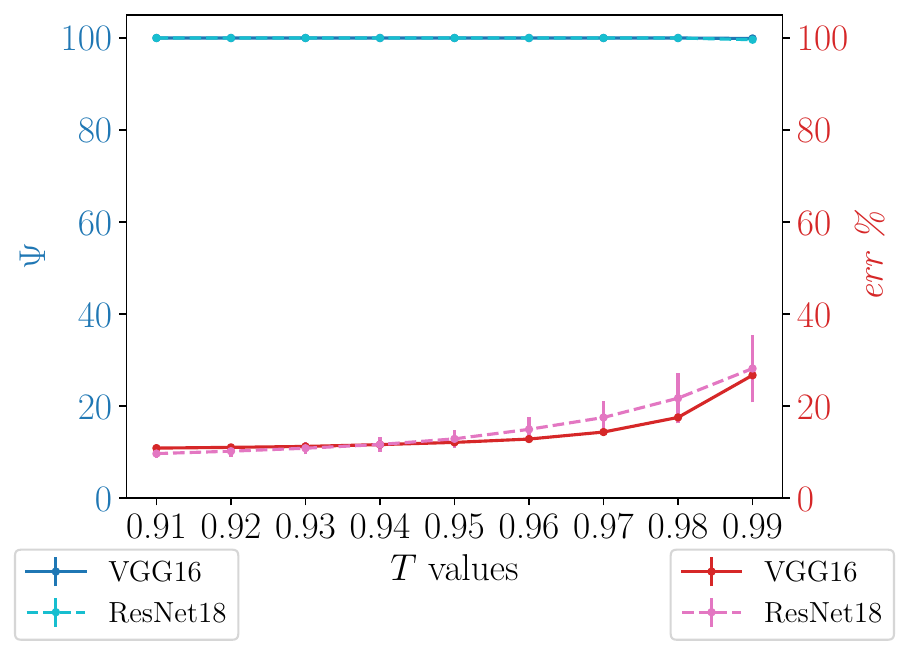}
\caption{Robustness against magnitude pruning. $\Psi$ is on the left axis in blue and the \textit{err \%} on the right axis in red.}
%\vspace{-10pt}
\label{fig:prun}
\end{subfigure}
~
\begin{subfigure}{0.49\textwidth}
\includegraphics[width=\textwidth]{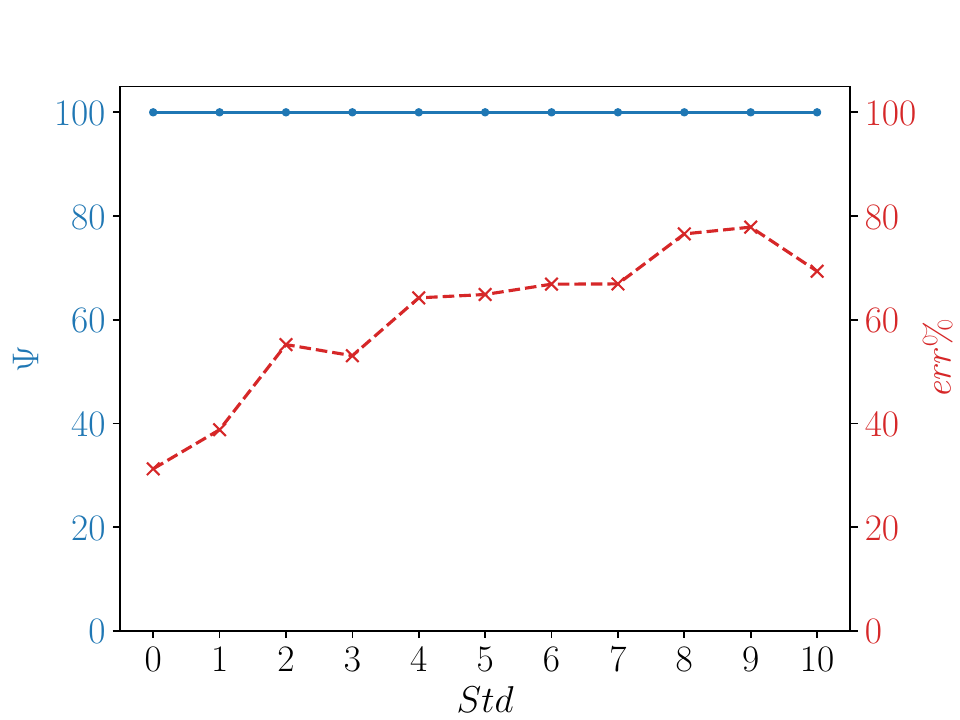}
\caption{Impact of the scalar product modification. $\Psi$ is on the left axis in blue and the \textit{err \%} on the right axis in red. }
\label{fig:citys2}
\end{subfigure}
\caption{Robustness analysis for VGG-16 and ResNet18 trained on CIFAR-10.}
\end{figure}

In this section, we propose some figures, showing robustness to Gaussian noise (Fig.~\ref{fig:addgauss}), robustness against quantization (Fig.~\ref{fig:quantiz}) and robustness against pruning (Fig.~\ref{fig:prun}).

Finally, we propose a plot showing the impact of modifying the scalar product on the model's behavior (integrity attack), in Fig.~\ref{fig:citys2}.

\section{Derivation of (9)}
\label{app:deriv1}
Let us evaluate the similarity score
\begin{equation}
    \label{eq2:1}
    S_C = \frac{\boldsymbol{w} \cdot \tilde{\boldsymbol{w}}}{\|\boldsymbol{w}\|_2\cdot \|\tilde{\boldsymbol{w}}\|_2}
\end{equation}
where we drop all the indices for abuse of notation. We can write $\tilde{\boldsymbol{w}}$ as 
\begin{equation}
    \tilde{\boldsymbol{w}} = \boldsymbol{w} + \hat{\boldsymbol{w}}
\end{equation}
where $\hat{\boldsymbol{w}}$ is some perturbation applied to $\boldsymbol{w}$ to get $\tilde{\boldsymbol{w}}$. We can expand \eqref{eq2:1}:
\begin{equation}
    S_C = \frac{\sum_i w_i^2 + \sum_i w_i \hat{w}_i}{\sqrt{\sum_i w_i^2} \cdot \sqrt{\sum_i (w_i + \hat{w}_i)^2} }\\
\end{equation}
Since we are looking for the conditions such that $S_C=1$, we need to look for the conditions such that
\begin{align*}
    \sum_i w_i^2 &+ \sum_i w_i \hat{w}_i = \sqrt{\sum_i w_i^2} \cdot \sqrt{\sum_i (w_i + \hat{w}_i)^2}\left(\sum_i w_i^2\right)^2 + \left(\sum_i w_i \hat{w}_i\right)^2 + 2\cdot  \left(\sum_i w_i^2\right) \cdot \left(\sum_i w_i \hat{w}_i\right)\\
    &= \left(\sum_i w_i^2\right) \cdot \left(\sum_i w_i^2 + \sum_i \hat{w}_i^2 + 2\sum_i w_i \hat{w}_i \right)\left(\sum_i w_i^2\right)^2 + \left(\sum_i w_i \hat{w}_i\right)^2 + 2\cdot  \left(\sum_i w_i^2\right) \cdot \left(\sum_i w_i \hat{w}_i\right)=\\
    &= \left(\sum_i w_i^2\right)^2 + \left(\sum_i w_i^2\right)\cdot \left(\sum_i \hat{w}_i^2\right)+ 2\cdot  \left(\sum_i w_i^2\right) \cdot \left(\sum_i w_i \hat{w}_i\right).\\
\end{align*}
By simplifying, we obtain
\begin{equation*}
    \left(\sum_{i}w_{i}\hat{w}_{i}\right)^2 = \left(\sum_i w_{i}^2 \right) \cdot \left(\sum_i \hat{w}_{i}^2 \right),
\end{equation*}
finding back~(10).

\section{Derivation of (13)}
\label{app:deriv2}
Let us assume the input follows a gaussian distribution, with mean $\boldsymbol{\mu}$ and covariance matrix $\Sigma$. We know that the post-synaptic potential, still follows a gaussian distribution, having
\begin{equation*}
    \mu_z = \sum_i w_i \mu_i
\end{equation*}
\begin{equation*}
    \sigma_z^2 = \sum_i w_i \left(w_i \Sigma_{ii} + 2\sum_{i'<i} w_{i'}\Sigma_{ii'}\right)
\end{equation*}
When introducing a perturbation $\boldsymbol{\hat{w}} = k\cdot \boldsymbol{w}$ having $k$ scalar, we know 
\begin{align*}
    \tilde{\mu}_z =& \mu_z + \sum_i \hat{w}_i \mu_i = \mu_z + \hat{\mu}_z = (1+k)\mu_z\\
%\end{equation*}
%\begin{align*}
    \tilde{\sigma}_z^2 =& \sum_i (1+k)w_i \left((1+k)w_i \Sigma_{ii} + 2\sum_{i'<i} (1+k)w_{i'}\Sigma_{ii'}\right)
    = (1+k) \sum_i w_i \left(w_i \Sigma_{ii} + k w_i \Sigma_{ii} + 2\sum_{i'<i} w_{i'}\Sigma_{ii'}+ k w_{i'}\Sigma_{ii'}\right)\\
    =& (1+k) \left\{\left[\sum_i w_i \left(w_i \Sigma_{ii} + 2\sum_{i'<i} w_{i'}\Sigma_{ii'}\right)\right] + k\left[\sum_i w_i \left(w_i \Sigma_{ii} + 2\sum_{i'<i} w_{i'}\Sigma_{ii'}\right)\right]\right\}= (1+k)^2 \sigma_z^2
\end{align*}
Hence, we can write the KL divergence
\begin{align*}
    D_{\text{KL}}(z || \tilde{z}) =& \log (1+k) + \frac{\sigma_z^2+k^2\mu_z^2}{2(1+k)^2 \sigma_z^2} - \frac{1}{2}\\
\end{align*}
\iffalse
\begin{align*}
    \tilde{\sigma}_z^2 =& \sum_i( w_i+\hat{w}_i) \left( (w_i+\hat{w}_i) \Sigma_{ii} + 2\sum_{i'<i} (w_{i'}+\hat{w}_{i'})\Sigma_{ii'}\right)\\
    =& \sum_i \left[ (w_i^2 +\hat{w}_i^2 + 2 w_i \hat{w}_i) \Sigma_{ii} +\right .\\
    +&\left . 2\sum_{i'<i} \left((w_{i'}w_i+\hat{w}_{i'}w_i +\hat{w}_{i}w_{i'}  + \hat{w}_{i}\hat{w}_{i'} + )\Sigma_{ii'}\right]\right]\\
    =& \sigma_z^2 + \sum_i \left[ (k^2 w_i^2 + 2 k w_i^2 ) \Sigma_{ii} +\right .\\
    +&\left . 2\sum_{i'<i} \left(k w_{i'}w_i +k w_{i}w_{i'}  + k^2 w_{i}w_{i'}\right )\Sigma_{ii'}\right]\\
    =& \sigma_z^2 + k \sum_i w_i \left[ (k + 2)w_i \Sigma_{ii} + 2\sum_{i'<i} (k + 2)w_{i'} \Sigma_{ii'}\right]\\
    =& \sigma_z^2 + \hat{\sigma}_z^2
\end{align*}
From this, we can write a KL-divergence between $z$ and $\tilde{z}$
\begin{align*}
    D_{\text{KL}}(z || \tilde{z}) =& \frac{1}{2}\left[ \log\left(\frac{\sigma_z^2 + \hat{\sigma}_z^2}{\sigma_z^2}\right) + \frac{\sigma_z^2 + (\mu_z - \hat{\mu}_z -\mu_z)^2}{\sigma_z^2 + \hat{\sigma}_z^2} -1\right]\\
    =&\frac{1}{2}\left[ \log\left(1+\frac{\hat{\sigma}_z^2}{\sigma_z^2}\right) + \frac{\sigma_z^2 +  \hat{\mu}_z^2}{\sigma_z^2 + \hat{\sigma}_z^2} -1\right]\\
\end{align*}
Under the assumption of having an activation such that $|\varphi(x)| \leq |x| \forall x \in \mathbb{R}$\enzo{to verify/mention the theorem}, we know that the above divergence upper-bounds $D_{\text{KL}}(y || \tilde{y})$. More specifically, for ReLU activations
\fi
%\end{document}

\section{Derivation of~(14)}
\label{app:derivklrelu}
In the specific case of employing a ReLU activation, assuming $\mu_z = 0$ we know that 
\begin{equation}
    \left\{
    \begin{array}{ll}
    D_{\text{KL}}(y || \tilde{y}) = 0 & z \leq 0\\
    D_{\text{KL}}(y || \tilde{y}) = D_{\text{KL}}(z || \tilde{z}) & z > 0
    \end{array}\right . .
\end{equation}
Hence, we can write the KL divergence as
\begin{align*}
    D_{\text{KL}}(y || \tilde{y}) =& \int_0^{+\infty} \frac{1}{\sigma_z\sqrt{2\pi}}e^{-\frac{x^2}{2\sigma_z^2}}(1+k)\sigma_z\sqrt{2\pi} e^{\frac{x^2}{2(1+k)^2\sigma_z^2}} dx=\frac{1}{\sigma_z\sqrt{2\pi}} \int_0^{+\infty} e^{-\frac{x^2}{2\sigma_z^2}}\left[ \log(1+k) - \frac{k^2 x^2}{2(1+k)^2\sigma_z^2} \right ] dx\\
    =&\frac{1}{\sigma_z\sqrt{2\pi}} \frac{\sigma_z^3 \sqrt{2\pi}\left[ 2(k + 1)^2\log(k+1) -k^2-2k \right ]}{4\sigma_z^2 (k+1)^2}=\frac{2(k+1)^2\log(k+1) - k(k+2)}{(k+1)^2}\nonumber
\end{align*}

\section{Additional results}
Here follow the detailed tables (all the numbers for all ranges) for all four types of modifications. For Gaussian noise addition, the main document presents 5 values $\Omega = [0,1,2,7,10]$ while Table~\ref{tab2:addgauss} presents all the results for  $\Omega \in [1,10]$. For fine-tuning, Table~\ref{tab2:finetuning} just add one the value compare to the main document =. For quantization, the main document presents 5 values  $B = [2,4,6,8,16]$ while Table~\ref{tab2:quantiz} presents all the results for $B\in [2; 16]$. For magnitude pruning, the main document presents 5 values $T = [0,91,95,98,99]\%$ while Table~\ref{tab2:pruning} presents all the results for $T \in [90; 99]\%$.

\begin{table}[!h]
\centering
\caption{Robustness to Gaussian noise addition.} \label{tab2:addgauss}
\resizebox{0.9\textwidth}{!}{
\begin{tabular}{ c c c c c c c c c c c}
 \toprule
    \multirow{2}{*}{$\boldsymbol{\Omega}$}&&\multicolumn{2}{c}{\bf CIFAR10}&\multicolumn{4}{c}{\bf ImageNet}&\bf Cityscapes& \bf COCO & \bf UVG \\
    & & VGG16 & ResNet18 & ResNet50 & ResNet101 & ViT-b-32 & MobileNetV3 &DeepLab-v3&YOLO-v5n & DVC\\
\midrule
 \multirow{2}{*}{$0$} & $\Psi (\uparrow)$ & 100\small{$\pm0$} & 100\small{$\pm0$} & 100 & 100 & 100 & 100 & 100 & 100 & 100 \\
 & \it metric$(\downarrow)$ &  \it 9.96$^\dag$\small{$\pm0.19$} & \it 7.03$^\dag$\small{$\pm0.28$} & \it 23.85$^\dag$ & \it 22.63$^\dag$ & \it 24.07$^\dag$  & \it 25.95$^\dag$ & \it 32.26$^\ddag$ & \it 48.70$^\star$ & \it 0.177$^\bullet$\\
\midrule
\multirow{2}{*}{$1$} & $\Psi (\uparrow)$ & 100\small{$\pm0$} & 100\small{$\pm0$} & 100 & 100 & 100 & 100 & 100 & 75.69 & 100\\
 & \it metric$(\downarrow)$ & \it 10.25$^\dag$\small{$\pm0.17$} & \it 7.30$^\dag$\small{$\pm0.08$} & \it 24.73$^\dag$    & \it 23.17$^\dag$ & \it 24.08$^\dag$ & \it 40.44$^\dag$ & \it 32.16$^\ddag$ & \it 79$^\star$ & \it 0.177$^\bullet$\\
\midrule
  \multirow{2}{*}{$2$} & $\Psi (\uparrow)$ & 100\small{$\pm0$} & 100\small{$\pm0$} & 99.56 & 99.26 & 100 & 100 & 100 & 57.65 & 100\\
 & \it metric$(\downarrow)$& \it 11.82$^\dag$\small{$\pm0.17$} & \it 9.13$^\dag$\small{$\pm0.59$} & \it 27.68$^\dag$  & \it 25.08$^\dag$ & \it 24.77$^\dag$ & \it 70.50$^\dag$ & \it 32.75$^\ddag$ & \it 82.10$^\star$ & \it 0.177$^\bullet$ \\
\midrule
  \multirow{2}{*}{$3$} & $\Psi (\uparrow)$ & 100\small{$\pm0$} & 100\small{$\pm0$} & 98.24 & 97.26 & 100 & 100 & 100 & 35.29 & 100\\
 & \it metric$(\downarrow)$& \it 15.20$^\dag$\small{$\pm0.40$} & \it 13.95$^\dag$\small{$\pm2.24$} & \it 33.15$^\dag$  & \it 28.60$^\dag$ & \it 26.63$^\dag$ & \it 88.83$^\dag$ & \it 34.06$^\ddag$ & \it 94.14$^\star$ & \it 0.177$^\bullet$ \\
\midrule
  \multirow{2}{*}{$4$} & $\Psi (\uparrow)$ & 100\small{$\pm0$} & 100\small{$\pm0$} & 91.89 & 91.21 & 100 & 100 & 100 & 24.31 & 100 \\
 & \it metric$(\downarrow)$& \it 20.39$^\dag$\small{$\pm0.88$} & \it 23.47$^\dag$\small{$\pm4.93$} & \it 42.09$^\dag$  & \it 33.95$^\dag$ & \it 30.02$^\dag$ & \it 95.25$^\dag$ & \it 35.73$^\ddag$ & \it 90.49$^\star$ & \it 0.178$^\bullet$ \\
\midrule
  \multirow{2}{*}{$5$} & $\Psi (\uparrow)$ & 100\small{$\pm0$} & 100\small{$\pm0$} & 77.34 & 76.12 & 100 & 99.84 & 92.19 & 16.08 & 100\\
 & \it metric$(\downarrow)$& \it 27.13$^\dag$\small{$\pm1.34$} & \it 36.88$^\dag$\small{$\pm6.91$} & \it 54.28$^\dag$  & \it 41.36$^\dag$ & \it 37.05$^\dag$ & \it 97.46$^\dag$ & \it 37.06$^\ddag$ & \it 98.45$^\star$ & \it 0.180$^\bullet$ \\
\midrule
  \multirow{2}{*}{$6$} & $\Psi (\uparrow)$ & 100\small{$\pm0$} & 100\small{$\pm0$} & 57.28 & 56.78 & 100 & 99.62 & 64.69 & 7.84 & 100\\
 & \it metric$(\downarrow)$& \it 34.32$^\dag$\small{$\pm1.68$} & \it 49.46$^\dag$\small{$\pm7.19$} & \it 66.97$^\dag$  & \it 50.47$^\dag$ & \it 48.25$^\dag$ & \it 98.45$^\dag$ & \it 37.44$^\ddag$ & \it 99.28$^\star$ & \it 0.182$^\bullet$ \\
\midrule
  \multirow{2}{*}{$7$} & $\Psi (\uparrow)$ & 99.88\small{$\pm0.12$} & 99.90\small{$\pm0.10$} & 57.28  & 39.45 & 99.64 & 85.31 & 41.02 & 8.24 & 100 \\
 & \it metric$(\downarrow)$& \it 41.27$^\dag$\small{$\pm2.11$}  & \it 99.55$^\dag$\small{$\pm6.30$} & \it 77.49$^\dag$ & \it 60.49$^\dag$ & \it 60.39$^\dag$ &\it 98.91$^\dag$ & \it 37.96$^\ddag$ & \it 99.62$^\star$ & \it 0.180$^\bullet$ \\
  \midrule
    \multirow{2}{*}{$8$} & $\Psi (\uparrow)$ & 99.47\small{$\pm0.46$} & 99.59\small{$\pm0.28$} & 27.49 & 26.86 & 100 & 71.72 & 26.17 & 5.49 & 100\\
 & \it metric$(\downarrow)$& \it 52.48$^\dag$\small{$\pm2.23$} & \it 71.63$^\dag$\small{$\pm5.35$} & \it 84.62$^\dag$  & \it 69.53$^\dag$ & \it 70.79$^\dag$ & \it 99.16$^\dag$ & \it 38.30$^\ddag$ & \it 99.67$^\star$ & \it 0.187$^\bullet$ \\
\midrule
  \multirow{2}{*}{$9$} & $\Psi (\uparrow)$ & 97.68\small{$\pm0.49$} & 98.10\small{$\pm0.53$} & 19.53 & 18.07 & 100 & 54.37 & 19.12 & 8.24 & 100 \\
 & \it metric$(\downarrow)$& \it 56.62$^\dag$\small{$\pm2.34$} & \it 75.18$^\dag$\small{$\pm5.14$} & \it 89.52$^\dag$  & \it 77.11$^\dag$ & \it 78.58$^\dag$ & \it 99.30$^\dag$ & \it 39.62$^\ddag$ & \it 99.94$^\star$ & \it 0.182$^\bullet$ \\
\midrule
\multirow{2}{*}{$10$} & $\Psi (\uparrow)$ & 93.50\small{$\pm0.98$} & 94.9\small{$\pm0.80$} & 12.93 & 12.74 & 99.22 & 43.05 & 12.89 & 5.49 & 100\\
 & \it metric$(\downarrow)$ & \it 56.62$^\dag$\small{$\pm2.31$} & \it 75.18$^\dag$\small{$\pm5.14$} & \it 92.65$^\dag$ & \it 83.04$^\dag$ &\it 83.99$^\dag$ &\it 99.41$^\dag$ & \it 39.74$^\ddag$ & \it 99.23$^\star$ & \it 0.182$^\bullet$\\
 
  \bottomrule
 \end{tabular}
}
\end{table}

\begin{table*}[b]
\centering
\caption{Robustness to fine-tuning.} \label{tab2:finetuning}
\resizebox{0.9\textwidth}{!}{
\begin{tabular}{ c c c c c c c c c c c}
 \toprule
    \multirow{2}{*}{$\boldsymbol{\Omega}$}&&\multicolumn{2}{c}{\bf CIFAR10}&\multicolumn{4}{c}{\bf ImageNet}&\bf Cityscapes& \bf COCO & \bf UVG \\
    & & VGG16 & ResNet18 & ResNet50 & ResNet101 & ViT-b-32 & MobileNetV3 &DeepLab-v3&YOLO-v5n & DVC\\
 \midrule
 \multirow{2}{*}{$2$} & $\Psi (\uparrow)$ & 100\small{$\pm0$} & 100\small{$\pm0$} & 100 & 100 & 100 & 100 & 100 & 100 & 100\\
  & \it metric$(\downarrow)$  & \it 9.97$^\dag$\small{$\pm0.20$} & \it 6.96$^\dag$\small{$\pm0.11$} & \it 23.35$^\dag$ & \it 22.54$^\dag$ & \it 24.08$^\dag$ & \it 26.60$^\dag$  &\it 34.85$^\ddag$ & \it 47.40$^\star$ & \it 0.184$^\bullet$ \\
 \midrule
 \multirow{2}{*}{$4$} & $\Psi (\uparrow)$ & 100\small{$\pm0$} & 100\small{$\pm0$} & 100 & 100 & 100 & 100 & 100 & 100 & 100\\
  & \it metric$(\downarrow)$  & \it 9.93$^\dag$\small{$\pm0.24$} & \it 8.12$^\dag$\small{$\pm0.34$} & \it 23.22$^\dag$ & \it 22.52$^\dag$ & \it 24.08$^\dag$ & \it 26.51$^\dag$  &\it 32.09$^\ddag$ & \it 47.20$^\star$ & \it 0.180$^\bullet$ \\
\midrule
  \multirow{2}{*}{$6$} & $\Psi (\uparrow)$ & 100\small{$\pm0$} & 100\small{$\pm0$} & 100 & 100 & 100 & 100 & 100 & 100 & 100\\
 & \it metric$(\downarrow)$ & \it 9.96$^\dag$\small{$\pm0.19$} & \it 6.98$^\dag$\small{$\pm0.15$}  & \it 23.23$^\dag$ & \it 22.57$^\dag$ & \it 24.08$^\dag$ & \it 26.41$^\dag$ &\it 31.58$^\ddag$ & \it 46.20$^\star$ & \it 0.179$^\bullet$ \\
\midrule
\multirow{2}{*}{$8$} & $\Psi (\uparrow)$ & 100\small{$\pm0$} & 100\small{$\pm0$} & 100 & 100 & 100 & 100 & 100 & 100 & 100\\
 & \it metric$(\downarrow)$ & \it 9.88$^\dag$\small{$\pm0.24$} & \it 7.01$^\dag$\small{$\pm0.19$}  & \it 23.21$^\dag$ & \it 22.54$^\dag$ & \it 24.07$^\dag$ & \it 26.37$^\dag$ &\it 30.35$^\ddag$ & \it 46.40$^\star$ & \it 0.179$^\bullet$ \\
\midrule
 \multirow{2}{*}{$10$} & $\Psi (\uparrow)$ & 100\small{$\pm0$} & 100\small{$\pm0$} & 100 & 100 & 100 & 100 & 100 & 100 & 100\\
 & \it metric$(\downarrow)$ & \it 9.89$^\dag$\small{$\pm0.16$} & \it 6.94$^\dag$\small{$\pm0.13$}   & \it 23.13$^\dag$ & \it 22.49$^\dag$ & \it 24.07$^\dag$ & \it 26.31$^\dag$ &\it 29.64$^\ddag$ & \it 46.00$^\star$ & \it 0.178$^\bullet$ \\

  \bottomrule
 \end{tabular}
}
\end{table*}

\begin{table*}[b]
\centering
\caption{Robustness to quantization.} \label{tab2:quantiz}
\resizebox{0.91\textwidth}{!}{
\begin{tabular}{ c c c c c c c c c c c}
 \toprule
    \multirow{2}{*}{$\boldsymbol{\Omega}$}&&\multicolumn{2}{c}{\bf CIFAR10}&\multicolumn{4}{c}{\bf ImageNet}&\bf Cityscapes& \bf COCO & \bf UVG \\
    & & VGG16 & ResNet18 & ResNet50 & ResNet101 & ViT-b-32 & MobileNetV3 &DeepLab-v3&YOLO-v5n & DVC\\
 \midrule
 \multirow{2}{*}{$16$} & $\Psi (\uparrow)$ & 100\small{$\pm0$} & 100\small{$\pm0$} & 100 & 100 & 100 & 100&  100 & 100 & 100\\
 & \it error $(\downarrow)$ & \it 9.96$^\dag$\small{$\pm0.19$} & \it 7.03$^\dag$\small{$\pm0.28$} &  \it 23.85$^\dag$ & \it 22.63$^\dag$ & \it 24.09$^\dag$ & \it 25.96$^\dag$  & \it 32.03$^\ddag$ & \it 48.70$^\star$ & \it 0.177$^\bullet$\\
 \midrule
  \multirow{2}{*}{$15$} & $\Psi (\uparrow)$ & 100\small{$\pm0$} & 100\small{$\pm0$} & 100 & 100 & 100 & 100&  100 & 100 & 100\\
 & \it error $(\downarrow)$ & \it 9.97$^\dag$\small{$\pm0.19$} & \it 7.04$^\dag$\small{$\pm0.28$} &  \it 23.86$^\dag$ & \it 22.62$^\dag$ & \it 24.09$^\dag$& \it 25.94$^\dag$  & \it 31.61$^\ddag$ & \it 48.70$^\star$ & \it 0.177$^\bullet$\\
 \midrule
  \multirow{2}{*}{$14$} & $\Psi (\uparrow)$ & 100\small{$\pm0$} & 100\small{$\pm0$} & 100 & 100 & 100 & 100&  100 & 100 & 100 \\
 & \it error $(\downarrow)$ & \it 9.96$^\dag$\small{$\pm0.19$} & \it 7.03$^\dag$\small{$\pm0.28$} &  \it 23.85$^\dag$ & \it 22.64$^\dag$ & \it 24.09$^\dag$ & \it 25.94$^\dag$  & \it 31.63$^\ddag$ & \it 48.70$^\star$ & \it 0.177$^\bullet$\\
 \midrule
  \multirow{2}{*}{$13$} & $\Psi (\uparrow)$ & 100\small{$\pm0$} & 100\small{$\pm0$} & 100 & 100 & 100 & 100 &  100 & 100 & 100\\
 & \it error $(\downarrow)$ & \it 9.96$^\dag$\small{$\pm0.19$} & \it 7.03$^\dag$\small{$\pm0.28$} &  \it 23.84$^\dag$ & \it 22.63$^\dag$ & \it 24.09$^\dag$ & \it 25.93$^\dag$  & \it 31.63$^\ddag$ & \it 48.70$^\star$ & \it 0.177$^\bullet$\\
 \midrule
  \multirow{2}{*}{$12$} & $\Psi (\uparrow)$ & 100\small{$\pm0$} & 100\small{$\pm0$} & 100 & 100 & 100 & 100&  100 & 100 & 100\\
 & \it error $(\downarrow)$ & \it 9.97$^\dag$\small{$\pm0.19$} & \it 7.04$^\dag$\small{$\pm0.28$} &  \it 23.84$^\dag$ & \it 22.66$^\dag$ & \it 24.10$^\dag$ & \it 25.95$^\dag$  & \it 31.73$^\ddag$ & \it 48.90$^\star$ & \it 0.177$^\bullet$\\
 \midrule
  \multirow{2}{*}{$11$} & $\Psi (\uparrow)$ & 100\small{$\pm0$} & 100\small{$\pm0$} & 100 & 100 & 100 & 100&  100 & 100 & 100\\
 & \it error $(\downarrow)$ & \it 9.98$^\dag$\small{$\pm0.18$} & \it 7.03$^\dag$\small{$\pm0.28$} &  \it 23.84$^\dag$ & \it 22.72$^\dag$ & \it 24.10$^\dag$ & \it 25.99$^\dag$  & \it 31.89$^\ddag$ & \it 48.90$^\star$ & \it 0.179$^\bullet$\\
 \midrule
   \multirow{2}{*}{$10$} & $\Psi (\uparrow)$ & 100\small{$\pm0$} & 100\small{$\pm0$} & 100 & 100 & 100 & 100&  100 & 100 & 100\\
 & \it error $(\downarrow)$ & \it 9.97$^\dag$\small{$\pm0.19$} & \it 7.04$^\dag$\small{$\pm0.28$} &  \it 23.84$^\dag$ & \it 22.72$^\dag$ & \it 24.10$^\dag$ & \it 25.99$^\dag$  & \it 31.26$^\ddag$ & \it 49.10$^\star$ & \it 0.178$^\bullet$\\
 \midrule
  \multirow{2}{*}{$9$} & $\Psi (\uparrow)$ & 100\small{$\pm0$} & 100\small{$\pm0$} & 100 & 100 & 100 & 100&  100 & 100 & 100\\
 & \it error $(\downarrow)$ & \it 9.97$^\dag$\small{$\pm0.18$} & \it 7.03$^\dag$\small{$\pm0.27$} &  \it 23.92$^\dag$ & \it 22.69$^\dag$ & \it 24.10$^\dag$ & \it 26.00$^\dag$  & \it 30.85$^\ddag$ & \it 47.60$^\star$ & \it 0.203$^\bullet$\\
 \midrule
\multirow{2}{*}{$8$} & $\Psi (\uparrow)$ & 100\small{$\pm0$} & 100\small{$\pm0$} & 100 & 100 & 100 & 100 & 100 & 100 & 98.44 \\
 & \it error $(\downarrow)$ & \it 9.97$^\dag$\small{$\pm0.20$} & \it 7.05$^\dag$\small{$\pm0.27$} & \it 24.03$^\dag$ & \it 23.11$^\dag$ & \it 24.10$^\dag$ & \it 26.37$^\dag$ & \it  31.84$^\ddag$ & \it 48.90$^\star$ & \it 0.338$^\bullet$\\
 \midrule
 \multirow{2}{*}{$7$} & $\Psi (\uparrow)$ & 100\small{$\pm0$} & 100\small{$\pm0$} & 100 & 100 & 100 & 100 & 100 & 100 & 100\\
 & \it error $(\downarrow)$ & \it 10.00$^\dag$\small{$\pm0.20$} & \it 7.06$^\dag$\small{$\pm0.29$} & \it 24.42$^\dag$ & \it 25.11$^\dag$ & \it 24.13$^\dag$ & \it 27.93$^\dag$ & \it  32.46$^\ddag$& \it 56.40$^\star$ & \it 2.20$^\bullet$\\
 \midrule
  \multirow{2}{*}{$6$} & $\Psi (\uparrow)$ & 100\small{$\pm0$} & 100\small{$\pm0$} & 100 & 100 & 100 & 100  & 100 & 100 & 98.44\\
 & \it error $(\downarrow)$ & \it 9.99$^\dag$\small{$\pm0.17$} & \it 7.14$^\dag$\small{$\pm0.27$} & \it 26.98$^\dag$ & \it 25.12$^\dag$ & \it 29.55$^\dag$ & \it 42.91$^\dag$ & \it 32.03$^\ddag$ & \it 89.10$^\star$ & \it 0.823$^\bullet$\\
 \midrule
  \multirow{2}{*}{$5$} & $\Psi (\uparrow)$ & 100\small{$\pm0$} & 100\small{$\pm0$} & 100 & 100 & 100 & 100 & 100 & 100 & 98.44\\
 & \it error $(\downarrow)$ & \it 10.13$^\dag$\small{$\pm0.28$} & \it 7.27$^\dag$\small{$\pm0.39$} & \it 97.19$^\dag$ & \it 99.17$^\dag$ & \it 75.83$^\dag$ & \it 94.82$^\dag$ & \it 39.40$^\ddag$ & \it 99.90$^\star$ & \it $\infty^\bullet$\\
 \midrule
 \multirow{2}{*}{$4$} & $\Psi (\uparrow)$ & 100\small{$\pm0$} & 100\small{$\pm0$} & 100 & 100 & 100 & 100 & 100 & 85.49 & 98.44\\
 & \it error $(\downarrow)$ & \it 10.77$^\dag$\small{$\pm0.28$} & \it 7.76$^\dag$\small{$\pm0.26$} & \it 99.91$^\dag$ & \it 99.90$^\dag$  & \it 96.81$^\dag$ & \it 99.91$^\dag$ & \it 47.28$^\ddag$ & \it 100$^\star$ & \it $\infty^\bullet$\\
 \midrule
  \multirow{2}{*}{$3$} & $\Psi (\uparrow)$ & 100\small{$\pm0$} & 100\small{$\pm0$} & 100 & 96.48 & 100 & 100 & 100 & 61.18 & 89.06\\
 & \it error $(\downarrow)$ & \it 15.07$^\dag$\small{$\pm0.83$} & \it 11.61$^\dag$\small{$\pm1.19$} & \it 99.90$^\dag$ & \it 99.90$^\dag$  & \it 96.71$^\dag$ & \it 99.89$^\dag$ & \it 89.93$^\ddag$ & \it 100$^\star$ & \it $\infty^\bullet$\\
 \midrule
 \multirow{2}{*}{$2$} &  $\Psi (\uparrow)$ & 100\small{$\pm0$} & 100\small{$\pm0$} & 31.54 & 9.91 & 100 & 100 & 100 & 56.47 & 98.44\\
 & \it error $(\downarrow)$ & \it 87.73$^\dag$\small{$\pm5.56$} & \it 88.20$^\dag$\small{$\pm2.77$} & \it 99.90$^\dag$ & \it 99.90$^\dag$ & \it 99.81$^\dag$ & \it 99.90$^\dag$ & \it 96.63$^\ddag$ & \it 100$^\star$ & \it $\infty^\bullet$\\

  \bottomrule
 \end{tabular}
}
\end{table*}

\begin{table*}[t]
\centering
\caption{Robustness to magnitude pruning.} \label{tab2:pruning}
\resizebox{0.91\textwidth}{!}{
\begin{tabular}{ c c c c c c c c c c c}
 \toprule
    \multirow{2}{*}{$\boldsymbol{\Omega}$}&&\multicolumn{2}{c}{\bf CIFAR10}&\multicolumn{4}{c}{\bf ImageNet}&\bf Cityscapes& \bf COCO & \bf UVG \\
    & & VGG16 & ResNet18 & ResNet50 & ResNet101 & ViT-b-32 & MobileNetV3 &DeepLab-v3&YOLO-v5n & DVC\\
 \midrule
   \multirow{2}{*}{$0$} & $\Psi (\uparrow)$ & 100\small{$\pm0$} & 100\small{$\pm0$} & 100 & 100 & 100 & 100 & 100 & 100 & 100\\
 & \it metric$(\downarrow)$ & \it 9.96$^\dag$\small{$\pm0.19$} & \it 7.03$^\dag$\small{$\pm0.28$}& \it 23.85$^\dag$ & \it 22.63$^\dag$ & \it 24.07$^\dag$  & \it 25.95$^\dag$ & \it 32.26$^\ddag$ & \it 48.70$^\star$ & \it 0.177$^\bullet$\\
 \midrule
 \multirow{2}{*}{$0.91$} & $\Psi (\uparrow)$ & 100\small{$\pm0$} & 100\small{$\pm0$} & 98.63 & 98.34 & 100 & 100 & 100 & 65.88 & 100\\
 & \it metric$(\downarrow)$ & \it 10.87$^\dag$\small{$\pm0.22$} & \it 9.67$^\dag$\small{$\pm0.95$} & \it 26.57$^\dag$ & \it 23.94$^\dag$ & \it 25.06$^\dag$ & \it 40.00$^\dag$ & \it 33.11$^\ddag$ & \it 50$^\star$ & \it 0.188$^\bullet$\\
 \midrule
  \multirow{2}{*}{$0.92$} & $\Psi (\uparrow)$ & 100\small{$\pm0$} & 100\small{$\pm0$} & 98.43 & 98.24 & 100 & 100 & 100 & 61.57 & 100\\
 & \it metric$(\downarrow)$ & \it 11.03$^\dag$\small{$\pm0.33$} & \it 10.19$^\dag$\small{$\pm1.17$} & \it 26.85$^\dag$ & \it 24.14$^\dag$ & \it 25.21$^\dag$ & \it 41.62$^\dag$ & \it 33.40$^\ddag$ & \it 50.10$^\star$ & \it 0.188$^\bullet$\\
 \midrule
  \multirow{2}{*}{$0.93$} & $\Psi (\uparrow)$ & 100\small{$\pm0$} & 100\small{$\pm0$} & 98.29 & 97.94 & 100 & 100 & 100 & 59.61 & 100\\
 & \it metric$(\downarrow)$ & \it 11.23$^\dag$\small{$\pm0.35$} & \it 10.84$^\dag$\small{$\pm1.31$} & \it 27.38$^\dag$ & \it 24.20$^\dag$ & \it 25.25$^\dag$ & \it 43.18$^\dag$ & \it 33.58$^\ddag$ & \it 50.20$^\star$ & \it 0.192$^\bullet$\\
 \midrule
  \multirow{2}{*}{$0.94$} & $\Psi (\uparrow)$ & 100\small{$\pm0$} & 100\small{$\pm0$} & 96.97 & 96.48 & 100 & 100 & 100 & 56.47 & 100\\
 & \it metric$(\downarrow)$ & \it 11.63$^\dag$\small{$\pm0.40$} & \it 11.65$^\dag$\small{$\pm1.56$} & \it 27.80$^\dag$ & \it 24.39$^\dag$ & \it 25.41$^\dag$ & \it 45.75$^\dag$ & \it 33.78$^\ddag$ & \it 50.80$^\star$ & \it 0.189$^\bullet$\\
 \midrule
\multirow{2}{*}{$0.95$} & $\Psi (\uparrow)$ & 100\small{$\pm0$} & 100\small{$\pm0$} & 97.61 & 97.12 & 100 & 100 & 100 & 51.76 & 100\\
 & \it metric$(\downarrow)$& \it 12.11$^\dag$\small{$\pm0.45$} & \it 12.88$^\dag$\small{$\pm1.90$} & \it 28.35$^\dag$ & \it 24.58$^\dag$ &\it 25.60$^\dag$ & \it 48.73$^\dag$ & \it 33.75$^\ddag$ & \it 51.10$^\star$ & \it 0.191$^\bullet$ \\
  \midrule
\multirow{2}{*}{$0.96$} & $\Psi (\uparrow)$ & 100\small{$\pm0$} & 100\small{$\pm0$} & 96.97 & 96.48 & 100 & 100 & 100 & 47.05 & 100\\
 & \it metric$(\downarrow)$& \it 13.84$^\dag$\small{$\pm0.51$} & \it 14.94$^\dag$\small{$\pm2.67$} & \it 29.02$^\dag$ & \it 24.89$^\dag$ &\it 25.74$^\dag$ & \it 52.49$^\dag$ & \it 33.98$^\ddag$ & \it 51.60$^\star$ & \it 0.192$^\bullet$ \\
   \midrule
\multirow{2}{*}{$0.97$} & $\Psi (\uparrow)$ & 100\small{$\pm0$} & 100\small{$\pm0$} & 96.09 & 94.92 & 100 & 100 & 100 & 40.78 & 100\\
 & \it metric$(\downarrow)$& \it 14.36$^\dag$\small{$\pm0.85$} & \it 17.52$^\dag$\small{$\pm3.62$} & \it 29.81$^\dag$ & \it 25.23$^\dag$ &\it 26.03$^\dag$ & \it 57.67$^\dag$ & \it 34.62$^\ddag$ & \it 51.90$^\star$ & \it 0.205$^\bullet$ \\
 \midrule
  \multirow{2}{*}{$0.98$} & $\Psi (\uparrow)$ & 100\small{$\pm0$} & 100\small{$\pm0$} & 93.12 & 91.94 & 99.83 & 97.91 & 99.61 & 36.47 & 100\\
 & \it metric$(\downarrow)$& \it 17.53$^\dag$\small{$\pm1.28$}   & \it 21.71$^\dag$\small{$\pm5.39$} & \it 30.88$^\dag$ & \it 25.65$^\dag$ & \it 26.03$^\dag$ & \it 63.48$^\dag$ & \it 34.47$^\ddag$ & \it 52.60$^\star$ & \it 0.198$^\bullet$\\
  \midrule
\multirow{2}{*}{$0.99$} & $\Psi (\uparrow)$ & 99.90\small{$\pm0.13$} & 99.63\small{$\pm0.31$} & 81.10 & 78.81 & 95.37 & 86.46 & 90.24 & 25.88 & 100\\
 & \it metric$(\downarrow)$& \it 26.71$^\dag$\small{$\pm2.84$}  & \it 28.16$^\dag$\small{$\pm7.35$}  & \it 32.62$^\dag$ & \it 25.98$^\dag$ & \it 26.63$^\dag$ & 76.22$^\dag$  & \it 36.43$^\ddag$ & \it 41.40$^\star$ & \it 0.219$^\bullet$\\

  \bottomrule
 \end{tabular}
}
\end{table*}